\documentclass[final,5p,times,twocolumn,authoryear]{elsarticle}

\usepackage{amsmath, amssymb, amsthm}
\usepackage{booktabs}
\usepackage{csquotes}
\usepackage{comment}
\usepackage{csquotes}
\usepackage{graphicx} % Required for inserting images
\usepackage{listings}
\usepackage[hidelinks]{hyperref}
\usepackage[dvipsnames]{xcolor}
\usepackage{balance}
\usepackage{soul}

\makeatletter
\apptocmd\@makecaption{\par}{}{%
  \errmessage{\noexpand\@makecaption could not be patched}%
}
\makeatother

\usepackage{inconsolata}

\definecolor{code_bg}{RGB}{239, 241, 245}
\definecolor{code_comment}{RGB}{79, 79, 105}
\definecolor{code_string}{RGB}{64, 160, 43}
\definecolor{code_keyword}{RGB}{210, 15, 57}
\definecolor{code_number}{RGB}{254, 100, 11}

\newcommand{\flox}[0]{\textsc{Flight}}

\newcommand{\coordinator}{\texttt{Coordinator}}
\newcommand{\aggr}{\texttt{Aggregator}}
\newcommand{\aggrs}{\texttt{Aggregators}}
\newcommand{\worker}{\texttt{Worker}}
\newcommand{\workers}{\texttt{Workers}}

\newcommand{\future}{\texttt{Future}}
\newcommand{\futures}{\texttt{Futures}}
\newcommand{\result}{\texttt{JobResult}}
\newcommand{\job}{\texttt{Job}}
\newcommand{\strategy}{\texttt{Strategy}}

\newcommand{\launcher}{\texttt{Launcher}}

\newcommand{\coordstrat}{\texttt{CoordinatorStrategy}}
\newcommand{\aggrstrat}{\texttt{AggregatorStrategy}}
\newcommand{\workerstrat}{\texttt{WorkerStrategy}}
\newcommand{\trainerstrat}{\texttt{TrainerStrategy}}

\newcommand{\params}[0]{\omega}

\newcommand{\Data}[0]{\mathcal{D}}
\newcommand{\loss}[0]{\ell}

\def\equationautorefname~#1\null{Eq.~(#1)\null}

\newcommand{\nathaniel}[1]{{\color{Blue} \textbf{Nathaniel:} \enquote{#1}}}

\newcommand{\matt}[1]{{\color{Maroon} \textbf{Matt:} \enquote{#1}}}

\newcommand{\kyle}[1]{{\color{red} \textbf{Kyle:} \enquote{#1}}}

\usepackage{pifont}% http://ctan.org/pkg/pifont
\newcommand{\cmark}{\ding{51}}%
\newcommand{\xmark}{\ding{55}}%
\newcommand{\greencell}{\cellcolor[HTML]{77E085}}
\newcommand{\yellowcell}{\cellcolor[HTML]{E8F15C}}
\newcommand{\redcell}{\cellcolor[HTML]{E09077}}
\newcommand{\ccell}{\greencell\cmark}
\newcommand{\xcell}{\redcell\xmark}

\newcommand{\nacell}{\yellowcell N/A}

\usepackage[column=0]{cellspace}
\usepackage{colortbl}
\usepackage[noadjust]{cite}

\journal{Future Generation Computer Systems}

\begin{document}

\begin{frontmatter}

%% Title, authors and addresses

%% use the tnoteref command within \title for footnotes;
%% use the tnotetext command for theassociated footnote;
%% use the fnref command within \author or \affiliation for footnotes;
%% use the fntext command for theassociated footnote;
%% use the corref command within \author for corresponding author footnotes;
%% use the cortext command for theassociated footnote;
%% use the ead command for the email address,
%% and the form \ead[url] for the home page:
%% \title{Title\tnoteref{label1}}
%% \tnotetext[label1]{}
%% \author{Name\corref{cor1}\fnref{label2}}
%% \ead{email address}
%% \ead[url]{home page}
%% \fntext[label2]{}
%% \cortext[cor1]{}
%% \affiliation{organization={},
%%            addressline={}, 
%%            city={},
%%            postcode={}, 
%%            state={},
%%            country={}}
%% \fntext[label3]{}

\title{
    \flox{}: A % Simple
    FaaS-Based Framework for Complex and Hierarchical Federated Learning 
    % at Scale
}

%% use optional labels to link authors explicitly to addresses:
%% \author[label1,label2]{}
%% \affiliation[label1]{organization={},
%%             addressline={},
%%             city={},
%%             postcode={},
%%             state={},
%%             country={}}
%%
%% \affiliation[label2]{organization={},
%%             addressline={},
%%             city={},
%%             postcode={},
%%             state={},
%%             country={}}

\author[uc,anl]{Nathaniel Hudson}
\author[uc,anl]{Valerie Hayot-Sasson}
\author[uc]{Yadu Babuji}
\author[uc]{Matt Baughman}
\author[uc]{J. Gregory Pauloski}
\author[anl]{Ryan Chard}
\author[uc,anl]{Ian Foster}
\author[uc,anl]{Kyle Chard}
\affiliation[uc]{%
    organization={Department of Computer Science, University of Chicago},
    addressline={5801 S Ellis Ave}, 
    city={Chicago},
    postcode={60637}, 
    state={IL},
    country={United States}
}
\affiliation[anl]{%
    organization={Data Science and Learning Division, Argonne National Laboratory},
    addressline={9700 S Cass Ave},
    city={Lemont},
    postcode={60439}, 
    state={IL},
    country={United States}
}

\begin{abstract}
    % \nathaniel{150 word limit.}
    Federated Learning (FL) is a decentralized machine learning paradigm where models are trained on distributed devices and are aggregated at a central server. Existing FL frameworks assume simple two-tier network topologies where end devices are directly connected to the aggregation server. While this is a practical mental model, it does not exploit the inherent topology of real-world %capture real-world connectivity of complex 
    distributed systems like the Internet-of-Things. We present \flox{}, a novel FL framework that supports complex hierarchical multi-tier topologies, asynchronous aggregation, and decouples the control plane from the data plane. 
    %\flox{} enables FL workflows spanning multiple hierarchies of aggregation to more closely resemble the natural underlying network topology and is able to scale to thousands of workers. 
    We compare the performance of \flox{} against Flower, a state-of-the-art FL framework. Our results show that \flox{} scales beyond Flower, supporting up to 2048 simultaneous devices, and reduces FL makespan across several models. 
    Finally, we show that \flox{}'s 
    hierarchical FL model can %features in simulation and 
    %can support hierarchical FL scenarios when deployed on a real-world distributed testbed to demonstrate the reduction of 
    reduce communication overheads by more than 60\%.
\end{abstract}

%%Graphical abstract
%\begin{graphicalabstract}
%\includegraphics{grabs}
%\end{graphicalabstract}

%%Research highlights
%\begin{highlights}
%\item Research highlight 1
%\item Research highlight 2
%\end{highlights}

\begin{keyword}
%% keywords here, in the form: keyword \sep keyword, up to a maximum of 6 keywords
    Federated Learning \sep 
    Hierarchical Federated Learning \sep 
    Function-as-a-Service \sep 
    Decentralized Systems \sep
    Edge Intelligence

%% PACS codes here, in the form: \PACS code \sep code

%% MSC codes here, in the form: \MSC code \sep code
%% or \MSC[2008] code \sep code (2000 is the default)

\end{keyword}

\end{frontmatter}

%\tableofcontents

%% \linenumbers

%% main text
\section{Introduction}
% The growing presence of decentralized systems, such as mobile edge computing~\citep{etsi_mec} and \textit{Internet-of-Things}~(IoT)~\citep{iot_survey1, iot_survey2} systems, has led to an explosive growth in the amount of data created, generated, sensed, and consumed.
Much of today's data is naturally distributed due to the growing ubiquity of systems like the \textit{Internet-of-Things}~(IoT)~\citep{iot_survey1, baccour2022pervasive} and \textit{Mobile Edge Computing}~\citep{etsi_mec}. 
Conventionally, training an Artificial Intelligence~(AI) model on distributed data required first transferring the data to a centralized computing system (e.g., high-performance computing cluster). However, in many scenarios this approach is intractable 
due to large data volumes, data transfer costs, and privacy concerns~\citep{ali2022federated}.
%costs for real-time data transmission. 
%Additionally, many decentralized systems have rigid privacy concerns that must also be considered~\citep{ali2022federated}.
% This is due to both the cost associated with simply transferring the data itself as well as the possible privacy concerns.
% Decentralized learning paradigms, like 
\textit{Federated Learning}~(FL)~\citep{fl_seminal_1, fl_seminal_2}, provides a potential solution as it implements a distributed training paradigm in which AI models can be trained  in a distributed fashion without needing to relocate data.
%necessary to perform this knowledge extraction across large, decentralized systems. 
% One such paradigm that has garnered much attention is 

\begin{figure}[t]
    \centering
    \includegraphics[width=\linewidth]{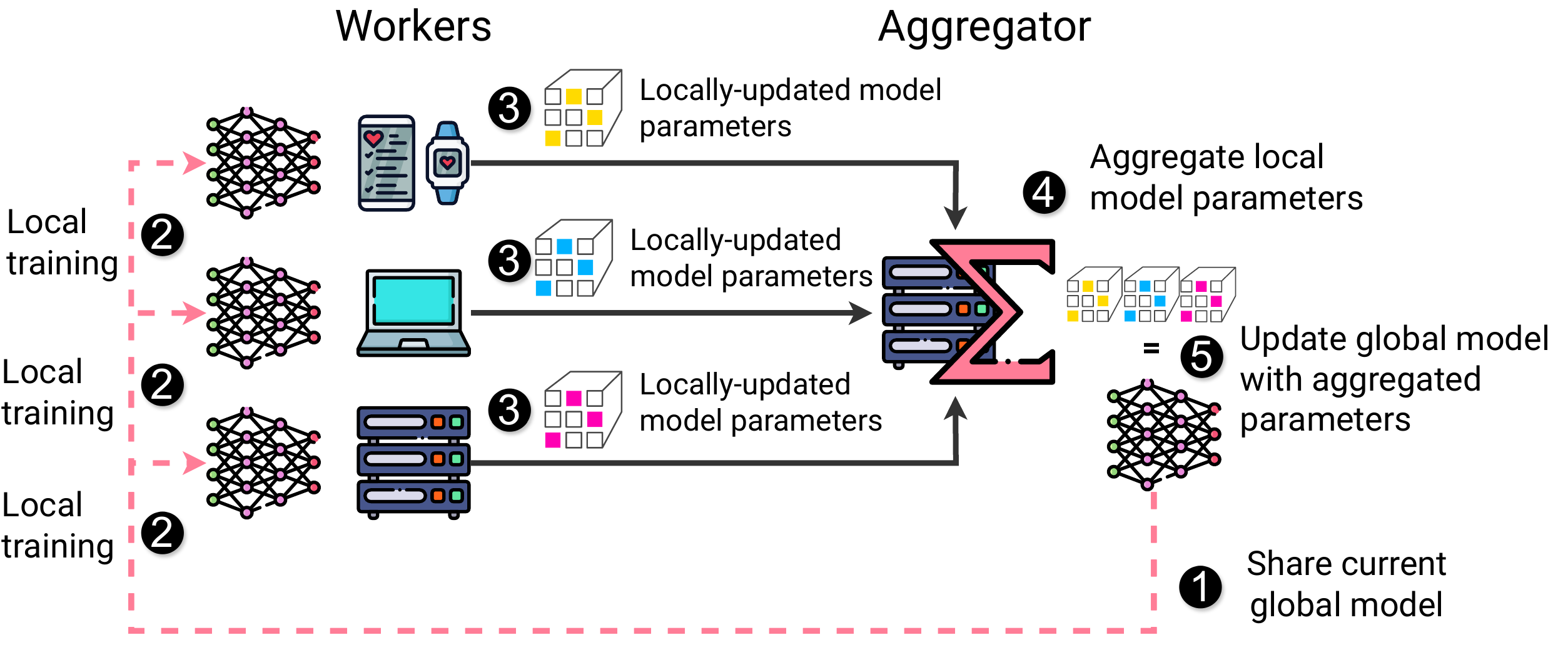}
    \caption{
        High-level view of a standard two-tier FL system and process.
        % \ian{Here and in next picture, the cloud symbols are presumably to imply that aggregators are hosted in the `cloud.' But that need not be the case?}
        % \nathaniel{That does not need to be the case. I can change it if you think it's necessary.} \kyle{would it be easy to replace one cloud with a server picture, e.g., from the top workers?}
        % \nathaniel{So, just put a server behind the aggregator? If so, I can do that very easily} \kyle{yea I think so} \kyle{Font size needs to be bigger too}
    }
    \label{fig:fl_example}
\end{figure}

% Unlike conventional approaches for training AI on decentralized data, FL does not require that raw (training) data  be moved to a central location. 
% Instead,
Unlike conventional deep learning, FL trains individual models directly where data reside 
%at the network edge 
(e.g., edge devices, IoT devices, mobile devices, and sensors). 
%, with the architecture of the model to be trained held constant across all end devices. 
A central location (e.g., server) is then tasked with aggregating (or averaging) locally-trained models rather than training a single model itself. \autoref{fig:fl_example} shows the general FL training process. 
\begin{comment}
As shown in \autoref{fig:fl_example}, the general cycle of a standard FL-driven training loop
is simple: %can be simply described as follows. 
%First, 
a central aggregation server initializes a global AI model with random weights; 
a copy of this global model is sent to each end device; each device performs local training using its own private data; 
devices then transfer the %locally-updated 
parameters of their local model to the aggregation server; 
the aggregation server aggregates the parameters received from the end devices and updates the global model accordingly; 
and then the loop repeats. 
% A visual of this loop is shown in \autoref{fig:fl_example}.
\end{comment}
Because no training data are communicated over the network in FL, it provides two key benefits:
\textit{(i)} reduced communication cost~\citep{hudson2022smart}, assuming the size of the model weights are less than the training data; 
and
\textit{(ii)} enhanced data privacy~\citep{fl_seminal_1}.

FL typically %takes place in a simple, 
assumes a two-tier system made up of a single \textit{aggregator} connected to a flat layer of \textit{workers} that perform local training (see \autoref{fig:fl_example}). 
While reasonable in many cases~\citep{patros2022rural}, this assumption has notable limitations. 
First, it prevents an FL process from taking advantage of the naturally hierarchical, often geographically clustered, scale-free 
topology of many networks (e.g., the Internet)~\citep{barabasi2003scale}.
%First, many networks (e.g., the Internet) have a scale-free topology, with naturally-occurring clusters~\citep{barabasi2003scale}. 
% The flat, two-tier model presented in \autoref{fig:fl_example} cannot leverage intermediate network hops between end devices and the aggregation server. 
% Thus, the natural hierarchical structure of the Internet is not taken advantage of to optimize the communication of the model parameters through the network. 
%---meaning network costs can be improved from standard, two-tier FL. 
Second, it ignores the geospatial relationship between the data distributions at end devices. 
%for certain applications. 
For example, consider a collection of smart homes in which smart meters collect, monitor, and predict energy consumption. 
%The end devices here are an individual smart home. 
The data collected from individual smart homes often  follows geospatial patterns
related to income levels~\citep{hudson2021framework}. 

%for example due to more affluent neighborhoods having different energy consumption patterns than impoverished neighborhoods~\citep{hudson2021framework}.

\begin{figure}[t]
    \centering
    \includegraphics[width=\linewidth]{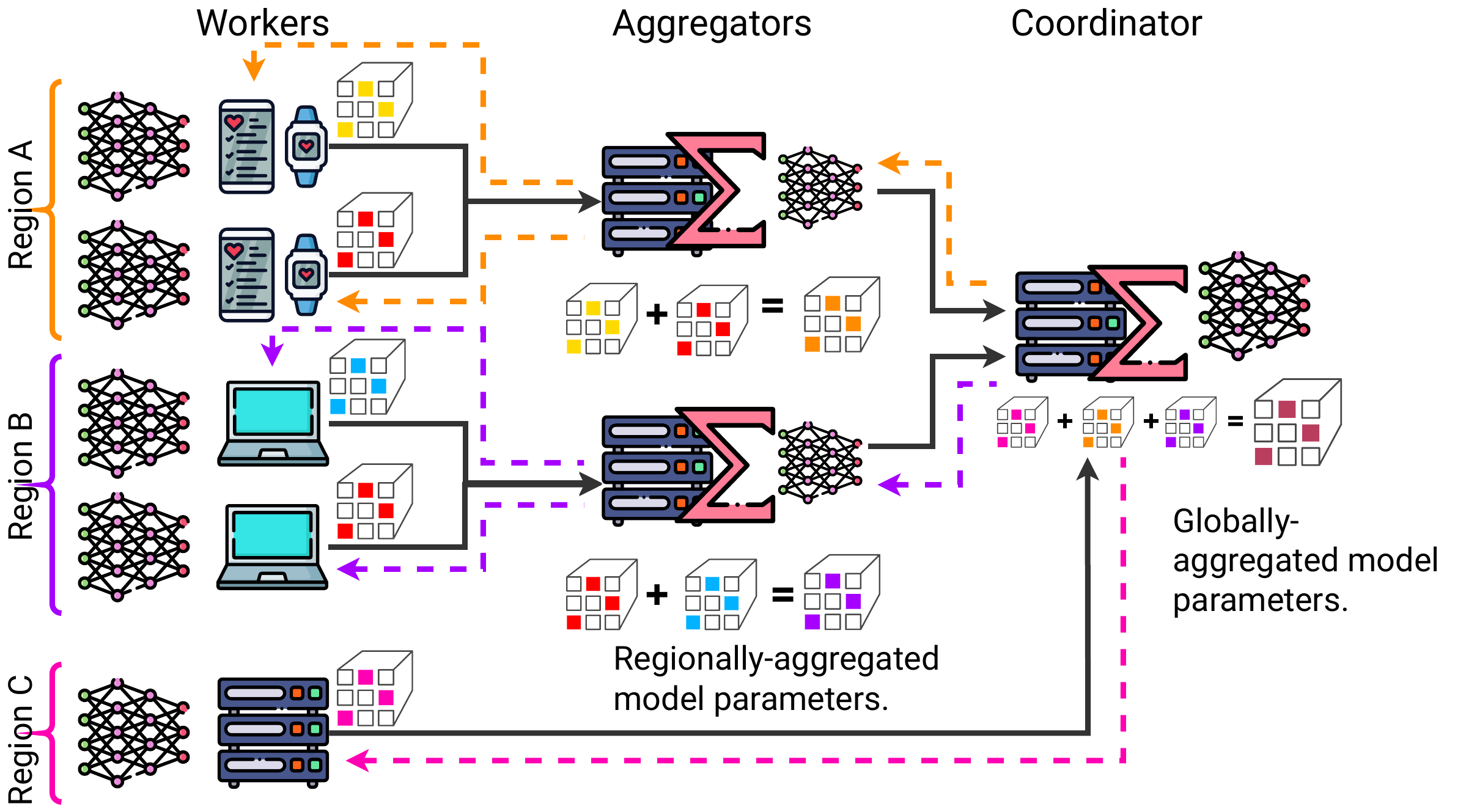}
    \caption{
        High-level view of a hierarchical FL system and process.  
        % \kyle{clouds \& fonts}
        % \ian{I notice that the black lines from the workers connect before the aggregator, while other do not. Is that significant?}
        % \nathaniel{No.}
    }
    \label{fig:hfl_example}
\end{figure}

\textit{Hierarchical Federated Learning}~(HFL)~\citep{async_hfl, hfl1, hfl2} %is a variant of FL that 
aims to address these problems. 
%as a solution to these problems is 
In HFL, the network used to share model parameter updates is multi-tier and hierarchical, 
and can include more than one aggregator in the network (see~\autoref{fig:hfl_example}). 
Intermediate aggregators can produce aggregated models that are more \textit{regional} in their context as they 
% The advantage of these intermediate regional models hosted by intermediate aggregators is 
ultimately depend on the data distribution of the workers from which local models are obtained. %and the network conditions.
HFL has been found to be particularly well-suited for use in remote environments with limited network connectivity due to its reduction of communication costs~\citep{almurshed2022adaptive,rana2022hierarchical}. 
%and reduction of single points of failure.
% \ian{And on the network characteristics?}
Despite recent innovation in the development of FL frameworks---e.g., Flower~\citep{flower}, FedML~\citep{fedml})---to the best of our knowledge there is no robust FL framework that natively supports complex HFL. Further, existing FL frameworks are often device-driven, an assumption that simplifies deployment, but does not scale to large, distributed systems. 
%when users deploy the system; however, for HFL, it is impractical to use device-driven FL due to the complexity of configuring many devices and aggregators and managing the FL hierarchy.

% To fill this gap, 
We present \textit{\flox{}}~(\textbf{F}ederated \textbf{L}earning \textbf{I}n \textbf{G}eneral \textbf{H}ierarchical \textbf{T}opologies), 
%which is---to the best of our knowledge---the first 
an open-source FL framework for implementing arbitrary hierarchies in distributed environments.%
\footnote{\url{https://github.com/globus-labs/flight}}  
\flox{} is the spiritual successor of our earlier FL framework known as FLoX~\citep{flox}.
Importantly, \flox{} supports both simulation and deployment on real devices across the computing continuum. To do so, \flox{} provides modular interfaces for control and data planes. In distributed environments, \flox{} can combine the Function-as-a-Service~(FaaS) paradigm (via Globus Compute) and ProxyStore to decouple control from data and enable flexible, efficient, and performant  
deployment.

\flox{} tackles a range of important distributed systems problems in FL. The primary contributions of our work are: % including deploying complex hierarchical FL models, decoupling control from data flow, supporting asynchronous aggregation,
%and scaling deployments to thousands of participating devices.

\begin{itemize}
    \item \flox{}, an open-source framework capable of defining and deploying hierarchical and asynchronous FL.
    % \item Enabling asynchronous FL across these hierarchies
    \item Methods to separate data and control flow in FL using robust compute and data-management frameworks.
    \item Comprehensive evaluation showing that \flox{} scales to thousands of concurrent workers, can reduce global data transfer in hierarchical topologies, can reduce training time using asynchronous FL, and can efficiently train in a distributed environment.
    % \item Enabling real-world FL by building on top of a state-of-the-art FaaS framework.
\end{itemize}
% On top of native support for HFL processes, \flox{} is highly modular---capable of enabling highly-customized research and deployment solutions for hierarchical and conventional FL processes.
% \flox{} also provides native support for asynchronous FL in two-tier FL systems.

% \kyle{We need to make sure to play up the systems side too. Some ideas - arbitrary hierarchy in a dist environment, separate control from data flow, async model, flexible exec model from HPC to globally distributed.}
% \ian{Be careful as to what is claimed as innovation, in the sense that this reads like its main (only?) distinctive feature is that it is an OS HFL system.}
% \kyle{Agreed. Other innovations are the ability to run in real environments. Support for PS to decouple data from control channel. Scalability beyond other frameworks (hopefully)}

The rest of this paper is as follows:
\autoref{sec:back} introduces hierarchical FL and terminology;
\autoref{sec:design} describes the \flox{} architecture;
\autoref{sec:strat} discusses how custom FL strategies can be implemented in \flox{};
\autoref{sec:eval} evaluates \flox{} in simulated and real environments;
% \autoref{sec:usecases} highlights some use cases implemented with \flox{};
% \autoref{sec:related} reviews related work;
	and
\autoref{sec:conc} concludes and discusses future directions.

\begin{comment}
\begin{enumerate*}[label=Section \arabic*]
	\setcounter{enumi}{1}
	\item provides a brief overview of FL and the language used to describe it in this paper,
	\item introduces and discusses the architecture underlying \flox{},
	\item discusses how custom FL strategies (or strategies) can be implemented in \flox{},
	\item presents the evaluation setup,
	\item highlights some implemented use cases with \flox{},
	\item showcases related work,
	and
	\item concludes the paper and discusses future directions.
\end{enumerate*}
\end{comment}

\section{Background \& Related Work}
\label{sec:back}

To better contextualize the contributions of \flox{}, we first formally define \textit{Hierarchical FL}~(HFL) and introduce HFL \textit{processes} (i.e., % are the abstraction that refer to the all the logical pieces needed in order to perform HFL. 
the various steps required to perform HFL, including model training,  transferring of parameters, and aggregation). We then survey existing FL frameworks and identify gaps that highlight the need for \flox{}.

%We will discuss the foundational details in this section and then briefly discuss existing FL frameworks before the introduction of \flox{}.

\subsection{Hierarchical Federated Learning Processes}
\label{sec:hfl_processes}

An HFL process is performed on a network topology made up of various connected devices. 
The topology of these devices are arranged as a tree (see \autoref{fig:hfl_example}).
The device at the root of this tree---the \textit{global aggregator} in \autoref{fig:hfl_example}---is responsible for coordinating the HFL process.
The first task of the global aggregator is to instantiate an ML model, in this case a deep neural network (DNN), called the \textit{global model}. 
The global model is initialized with random model parameters (or weights) $\params_{t=0}$ where $t$ denotes the current \textit{round} (i.e., $t=0$ is the initialization round).
% Upon being initialized, 
A copy of the global model is sent by the global aggregator to a worker deployed on an end device in the network topology. 
%that own data of interest to train on.
We refer to the copies of the global model now hosted by workers as \textit{local models}.
Workers then locally train their copy of the model on their local data. Local model parameters $\params_{t+1}^{k}$ for worker $k$ are updated according to \autoref{eq:local_training}:

\begin{equation}
    \params_{t+1}^{k} = \params_{t} - \eta \nabla \loss(\params_{t}, \Data_{k})
\label{eq:local_training}
\end{equation}

\noindent
where $\eta$ is the learning rate hyperparameter, $\loss(\params_{t}, \Data_{k})$ is the loss from using the global model parameters on the device's local dataset $\Data_{k}$, and $\nabla\loss(\cdot)$ is the gradient from the loss.

{\setlength{\tabcolsep}{6.9pt}
\begin{table*}
\centering
\scriptsize  % <-- Done to fit within columns
% \footnotesize{}
\begin{tabular}{rcccccccccc}
\toprule
	& \multicolumn{2}{c}{Hierarchies}
    & \multicolumn{3}{c}{Modularity} 
	& \multicolumn{3}{c}{Deployment}
    & \multicolumn{2}{c}{Control}
	\\
    \cmidrule(lr){2-3}
	\cmidrule(lr){4-6}
	\cmidrule(lr){7-9}
    \cmidrule(lr){10-11}
    % \\
    & \multicolumn{5}{c}{}
    & \multicolumn{2}{c}{Simulation}
    & \multicolumn{3}{c}{}
    \\
    \cmidrule(lr){7-8}
    
	Framework
    %Hierarchies
    & Complex
    & Simple
        
	% Modularity
	& Device Selection
	& Sync Aggr
	& Async Aggr
 
	% Deployment
	& Single-Node
	& Multi-Node 
	& Remote  % On-Device
    & Coord
    & Decoupled
	\\

\midrule

    LEAF~\citep{leaf}     
        & \xcell & \xcell
        & \xcell & \xcell & \xcell
        & \ccell & \xcell & \xcell 
        & \nacell &\nacell 
        \\

    TFF~\citep{tff} 
        & \xcell & \xcell
        & \xcell & \ccell & \xcell
        & \ccell & \ccell & \xcell 
        & \nacell  &\nacell 
	\\
 
    OpenFL~\citep{openfl}
        & \xcell & \xcell
        & \xcell & \ccell & \xcell
        & \ccell & \xcell & \xcell 
        & \nacell &\nacell 
        \\

    FLoX~\citep{flox}
        & \xcell & \xcell
        & \xcell & \ccell & \xcell
        & \ccell & \ccell & \ccell 
        & \ccell &\xcell 
        \\

    APPFL~\citep{appfl} 
        & \xcell & \xcell 
        & \xcell & \ccell & \ccell
        & \ccell & \ccell & \ccell 
        & \ccell &\xcell 
        \\
	
    Flower~\citep{flower} 
        & \xcell & \xcell
        & \ccell & \ccell & \xcell
        & \ccell & \ccell & \ccell 
        & \xcell &\xcell 
        \\

    FedScale~\citep{fedscale} 
        & \xcell & \xcell
        & \ccell & \ccell & \ccell
        & \ccell & \ccell & \ccell 
        & \ccell &\xcell 
        \\

    FedML~\citep{fedml}
        & \xcell & \ccell
        & \xcell & \ccell & \xcell
        & \ccell & \ccell & \ccell 
        & \ccell &\xcell 
        \\
    
    \textbf{\flox{} (ours)} 
        & \ccell & \ccell 
        & \ccell & \ccell & \ccell 
        & \ccell & \ccell & \ccell 
        & \ccell & \ccell
        \\
        
\bottomrule
\end{tabular}
\caption{%
    Overview of federated learning frameworks. % across several dimensions. 
    \textbf{Hierarchies} captures the topologies supported, from simple two- or three-tier to more complex topologies. \textbf{Modularity} captures the extensibility of the framework in terms of selecting participating devices and the use of synchronous or asynchronous aggregation. \textbf{Deployment} captures the scenarios supported from single-node to multi-node simulation and real deployment on devices. \textbf{Control} captures how the FL framework coordinates the FL process and if data and control planes are decoupled. \enquote{N/A} denotes that control coordination and data decoupling are not applicable in frameworks that do not support remote deployment.
}
\label{tab:frameworks}
\end{table*}
}

In standard two-tier FL, when the workers finish training their local models they send their locally-updated model parameters back to the global aggregator. In HFL processes,
%this is not necessarily true for all end devices (though it could be the case for some). Since the topologies we consider for HFL processes are organized as a tree, the 
workers instead send their locally-updated model parameters to their parent in the topology. This parent will either be an intermediate aggregator or the global aggregator. The intermediate aggregators in the tree are responsible for aggregating the model parameters returned by their topological children---irrespective of whether their children are leaves (workers) or \textit{other} intermediate aggregators. 
Various strategies can be used to aggregate the local models (see \autoref{sec:strat}).The simplest approach is to compute a simple average over the returned model parameters (see \autoref{eq:simple_avg}).
%This can
%\footnote{Note that many averaging methods exist for FL for a litany of different problems and scenarios---which we discuss later in \autoref{sec:strat}.} 
% be achieved using a simple average over the returned model parameters (see \autoref{eq:simple_avg}) or by using another FL strategy (see \autoref{sec:strat}).

\begin{equation}
    \params_{t+1}^{k} \triangleq 
        \frac{1}{\text{children}(k)}
        \sum_{k'\in\text{children}(k)} 
            \params_{t+1}^{k'}
\label{eq:simple_avg}
\end{equation}

Like the intermediate aggregators, the global aggregator will also collect some set of returned model parameters from its immediate children and then perform aggregation. 
The result of this aggregation is then used to update the global model. Once the global model is updated, the global aggregator might perform other administrative tasks (e.g., testing the global model, checkpointing) before launching a new round of local model training on the end devices.
%and aggregation on the intermediate aggregators and itself. 
% What separates HFL processes from standard, two-tier FL processes is the inclusion of the intermediate aggregators. 
% Two-tier FL simply considers the end devices that perform local training and that one global aggregator is responsible for aggregating all locally-updated parameters. 
% In HFL, we instead of a hierarchy of how model parameters are communicated back up to the root device to update the global model. 
Because we consider the intermediate aggregators as simply returning their own averaged parameters to their parents, we consider a two-tier FL process as a special case of HFL. 
% This mindset shaped much of the design of our proposed \flox{} framework, which we present in \autoref{sec:design}.

%%%%%%%%%%%%%%%%%%%%%%%%%%%%%%%%%%%%%%%%%%%%%%%%
%%%%%%%%%%%%%%%%%%%%%%%%%%%%%%%%%%%%%%%%%%%%%%%%

% \kyle{I wonder if we should introduce aggregation here too. E.g., methods for aggregation.}
% \nathaniel{Do you mean just briefly highlight some of the other algorithms from the literature, e.g., \texttt{FedAvg} and \texttt{FedProx}?}
The design of aggregation strategies is an active area of research in the FL community. The simplest strategy, \texttt{FedSGD}, performs a simple average, similar to \autoref{eq:simple_avg}, over returned model parameters in a two-tier setting.
McMahan et al.\ proposed the alternative \texttt{FedAvg} strategy~\citep{fl_seminal_1}, which generalizes \texttt{FedSGD} and is often seen as the standard aggregation strategy in FL. 
% This aggregation technique 
\texttt{FedAvg} foregoes a simple average and instead uses a weighted averaging method based on the amount of data at each end device. 
This can be defined as follows:
\begin{equation}
    \params_{t+1} \triangleq \sum_{k=1}^{K} \frac{n_k}{n} \params_{t+1}^{k}.
\label{eq:fedavg}
\end{equation}
The goal of \texttt{FedAvg} is to make the aggregation less sensitive to imbalanced data distributions.
Further, Li et al. later proposed \texttt{FedProx}~\citep{fedprox} which generalizes \texttt{FedAvg} with even greater emphasis on data heterogeneity. 
%This is done by adding a proximal term, defined as the norm between $\params_{t}$ and $\params_{t+1}^{k}$, before backpropagation during local training on the workers.
% 
Finally, alternative aggregation strategies exist for \textit{Asynchronous Federated Learning}~(AFL)~\citep{fedasync} to incorporate individual local model updates as they arrive. A simple approach is: % for AFL can be defined as

\begin{equation}
    \params_{t+1} = \beta \cdot \params_{t} + (1-\beta) \cdot \params_{t_{k'}}^{k}
\label{eq:async_avg}
\end{equation}

\noindent
where $t_{k'}$ is the time-step of the most recent update from worker~$k'$ and $\beta\in(0,1)$ is the step size.

\subsection{Federated Learning Frameworks}
\label{sec:existing_frameworks}
% \kyle{How much space do we have left? Would be good to expand this summary of the different FL systems if we have space?}
Here, we briefly survey existing FL frameworks, with prominent frameworks
%and current works and their capabilities 
summarized in~\autoref{tab:frameworks}.  
%Broadly speaking, these frameworks and tools can partitioned based in their areas of focus and corresponding strengths. We compare frameworks along four dimensions: hierarchy support, modularity, modes of deployment, and modes of control.

% \kyle{Move to textbf for each and explain what the otpions are} Within each of these primary areas, we break down support to include simple and complex hierarchies, the ability to intelligently select workers, built in methods for synchronous and asynchronous aggregation, and the ability to both simulate and deploy an FL system. Developing a framework that does all of these well is difficult as there are tradeoffs presented by incorporating very diverse use cases and functionality. Following, most frameworks try to specialize.

%\textbf{FL Frameworks}
% 
%These frameworks present unified standards for FL experiments, leaving users to implement support for distributed execution and to manage infrastructure. % and backend processes.

TensorFlow Federated (TFF)~\citep{tff} and LEAF~\citep{leaf}, two of the first FL frameworks, focus on training models using on-premise simulations. 
% 
% {\color{blue}
TFF is developed and maintained by Google and is meant to simulate FL processes and the underlying statistical qualities of federated datasets. 
Currently, it provides a lot of foundational abstractions for federated computations and mathematics; it is limited in its use for real-world FL use cases because it is moreso a simulation framework.
LEAF is a simple benchmarking framework for different FL scenarios using TensorFlow. It is not designed to be modular framework that enables rapid development of novel FL algorithms (e.g., aggregation algorithms). Additionally, it only provides the standard \texttt{FedAvg} algorithm for FL processes.
% }

%with limited support to incorporate distributed resources.
While FL was initially developed for communication efficiency~\citep{fl_seminal_1}, its ability to operate across sites enables privacy preservation when data cannot be moved from a device (e.g., medical datasets).
% \ian{Might one say that PP is often an important concern? One could also use FL for performance reasons alone? (or is that not really the case.)}
%However, substantial works have demonstrated the possibility of data leakage from model parameters in FL systems 
% \ian{perhaps relevant: \citep{ren2022grnn,jin2021cafe}} 
%using more sophisticated attacks~\citep{ren2022grnn, jin2021cafe}. While many of these works rely on limiting assumptions, such as that no more than eight data samples are used for a batch update~\citep{zhu2019deep, geiping2020inverting}, other works have shown that data distributions can be leaked without these assumptions~\citep{rajani2023adversarial}.
% 
% {\color{blue}
Some FL frameworks are more pointedly designed around the privacy-preservation benefits of FL.
APPFL~\citep{appfl} is a framework developed and maintained by a team at Argonne National Laboratory that performs cross-silo FL. APPFL is designed to be a platform that requires little technical expertise to use, including a graphical user interface to lessen the burden of entry. 
Similar to APPFL, SubstraFL~\citep{substra} is FL framework designed to enable medical research on naturally decentralized healthcare data.
PySyft~\citep{pysyft} is a general framework that provides algorithms for private deep learning. Though not necessarily a framework specifically made for FL, it provides out-of-the-box support for common privacy-preserving algorithms including differential privacy and homomorphic encryption.
% }
% 
%Such techniques include homomorphic encryption~\citep{hardy2017private}, data perturbation and anonymization~\citep{yin2021comprehensive}, and differential privacy~\citep{wei2020federated}.

Frameworks such as FedLess~\citep{fedless}, Flower~\citep{flower}, FedScale~\citep{fedscale},  and OpenFed~\citep{openfed} accommodate FL training over distributed resources and provide interfaces to customize the training process. These frameworks enable larger experiments as well as deployment on devices.  Additionally, frameworks like $\lambda$-FL~\citep{lambda_fl}, XFL~\citep{wang2023xfl}, and Parrot~\citep{parrot} offer simple user interfaces, % and customization, 
but lack support for broad deployments. 

The popular Flower~\citep{flower} FL framework supports a wide range of environments for both simulated and real world experiments. For deployment, Flower uses gRPC for communication. 
%and supports a very diverse set of physical hardware and platforms. 
Flower's experiments are client-driven but experimentally configured at the server. In this way, Flower represents a traditional client-server model where each is independently configured and waits on the other. 
%This model can simplify deployment as each client is self-contained and can be deployed
%both in batches or across distributed devices 
%and updated on the device side without changing the aggregation strategy semantics on the server. 
%However, to maintain this model, Flower does not function as modularly as would be necessary to support the full range of FL processes, particularly hierarchical experiments.

% FedML is the only mainstream framework that we know of that supports hierarchical federated learning. % directly within the framework. 
% It includes a wide suite of libraries to enable FL simulations, on-device experimentation, and cross-silo implementations. However, it supports only three-tier hierarchies (i.e., a central aggregator, sub-aggregators, and workers), limiting application  to cross-silo environments. 
%instead of more complex sensor networks in austere devices with limited networks.

% \kyle{add text to link the following subsections to the table. maybe also make them bold not textbfs? }

\textbf{Hierarchies:}
%As devices become increasingly distributed and models become larger, hierarchical structures of federated learning experiments become more natural as they increase communication efficiency and take advantage of naturally occurring network structures. 
We know of no other FL framework that, like \flox{}, supports HFL across hierarchical device networks in which a global aggregator may be connected to worker nodes by multiple intermediate aggregators in a tree topology.
%\flox{} allows for general HFL across such topologies, subject primarily to the constraint that a topology must form a tree with workers as leaves and the global aggregator as the root.
% (
One partial exception is FedML by~\citet{fedml}, which supports HFL in networks that link a global aggregator with multiple multi-GPU workers, with model training performed across GPUs on each individual worker followed by worker-localized aggregation before worker-aggregated parameters are returned to the global aggregator.
While useful for cross-silo FL, this feature is not sufficient for more sophisticated decentralized systems, such as Internet-of-Things, sensor networks, and mobile edge computing.
% )

\begin{comment}
Few FL frameworks enable use of hierarchical topologies with the notable exception of FedML~\citep{fedml,hier_fedml}. FedML implements a model that allows for three-tier hierarchical FL as a means of cross-silo learning. Each silo can have multiple devices, each of which runs an FL task, and an aggregator that is responsible for forwarding grouped results to a global aggregator. This is a natural expression of hierarchical FL but does not allow for more complex hierarchies found in practice (e.g., sensor networks).
%where there may be more than one level of hierarchies or where the topologies may be unbalanced.
\kyle{Describe explicitly  for each how FLIGHT is different}
\end{comment}

\textbf{Modularity:}
The ability to rapidly change FL strategies is crucial for experiments and real-world deployments (e.g., where some devices may be intermittently online).
%rapid experimentation and flexibility of design. 
Device selection allows for devices to be sampled in a training round, various strategies can be applied (e.g., based on data distribution, training time, or previous impact on global model). Flower provides extensible interfaces for this purpose.
%The key areas this modularity applies is to change out aggregation strategies and to perform worker selection. 
% Regarding aggregation, 

Most frameworks are designed to support different synchronous aggregation strategies~\citep{tff,openfl,flox,flower,fedml}; however,  
%These are the most common types of experiments and innovations when it comes to FL research. 
there are a handful of frameworks, such as APPFL~\citep{appfl} and FedScale~\citep{fedscale}, that support asynchronous aggregation. 
% \kyle{Way this is worded + cites is confusing. I think we should name the systems that do provide the feature by name here} 
This mode of aggregation is necessary in environments with highly heterogeneous or unreliable compute resources, allowing results to be incorporated as they are returned. 

\textbf{Deployment:}
% \kyle{More motivation here that many FL frameworks are only able to be used in toy environments.}
The vast majority of FL framework support single-node simulation and % used to conduct FL experiments. Most 
most now also support deploying simulations across several compute nodes and deploying on devices, with some notable exceptions~\citep{tff, openfl, leaf}. 
Most frameworks adopt a simple multi-processing approach for single node experiments. 
% {\color{blue}
Support for simulating FL is a necessary feature for a FL framework for simple debugging before remote deployment.
In addition to simple debugging, it is necessary for FL frameworks to enable simulation of FL processes for novel research in the field of FL. Not all FL researchers are necessarily interested in deploying on remote devices. For instance, FL researchers interested in developing novel privacy-preserving FL algorithms will only need to simulate FL to analyze the trade-offs of their proposed algorithms.
However, an ideal FL framework will have support for both simulations and real-world deployment and interoperable switching between these two modes.
Most existing FL frameworks 
% }
primarily use client-server or RPC models for multi-node and on-device deployment. 
Notable exceptions include FLoX and APPFL, which can use the FaaS-based Globus Compute platform to deploy training operations, and TFF, which uses Kubernetes.

% Flexibility in experimental settings is important but also requires the ability to extend these experiments to real-world environments. Nearly all frameworks we are aware of include the ability to do local, single node simulation. Simple scaling to multi-node simulation is slightly less common with several frameworks not supporting it or on-device deployment natively~\citep{tff,openfl} with LEAF~\citep{leaf} only lacking support for on-device deployments.

\textbf{Control:}
FL frameworks take different approaches to managing the FL process. 
% \matt{
\enquote{Coord} in \autoref{tab:frameworks} refers to whether the process is driven by the topmost node (i.e., the \enquote{Coordinator} in \autoref{fig:topologies}) rather than being driven the bottom-most nodes (i.e., the \enquote{Workers} in \autoref{fig:topologies}). 
The former also suggests that you can change model configuration and training hyperparameters from a centralized location.% (as opposed to altering the config locally as is needed in flower), 
\enquote{Decoupled} refers to whether the control plane is decoupled from the data plane, meaning that the communication of logical components (i.e., code for jobs) and data are handled separately.
% }
% Controlling deployments is essential to experimentation in the real-world, with two elements being most necessary---coordination of the FL process and decoupled data movement. For coordination, 
It is important to be able to easily reconfigure the FL process, including aggregation topologies, worker workloads, experiment parameters, and the model being used. Most FL frameworks---such as FLoX~\citep{flox}, APPFL~\citep{appfl}, FedScale~\citep{fedscale}, and FedML~\citep{fedml}---support such configurations; however device-driven approaches~\citep{flower} make reconfiguration challenging as device clients must be restarted to change parameters.
%by default uses a model of client configuration where different experiments can be run on the aggregation side but the client server will need to be reconfigured and restarted to change experimental parameters. 
This contrasts with other frameworks where new models and parameters can be pushed from the coordinator to reinitialize an experiment~\citep{flox}. 
% This rapidly increases the rate of iteration and centralizes control which can minimize errors of manual reconfiguration.

As we consider more sophisticated FL topologies, 
%it is no longer feasible to treat control and data in the same way. 
decoupling the data and control plane 
is essential for efficient and scalable deployment. %, and may also aid security. 
%ommunication patterns and ensuring experimental errors do not affect the primary communication and control of the clients. 
To the best of our knowledge, no existing framework separates control from data, nor do they provide a robust, standardized data communication method that is decoupled from the control plane.
Finally, FL frameworks take different approaches to communication with various assumptions regarding connectivity.
%the communication layer itself, each framework makes assumptions around permissions,
% inbound network access, and network reliability. 
For example, frameworks relying on gRPC assume inbound network access~\citep{appfl,flower} while others require SSH connections between devices~\citep{fedscale}. 

\section{\flox{}: Design \& Implementation}
\label{sec:design}

\begin{figure}[t]
    \centering
    \includegraphics[width=\linewidth]{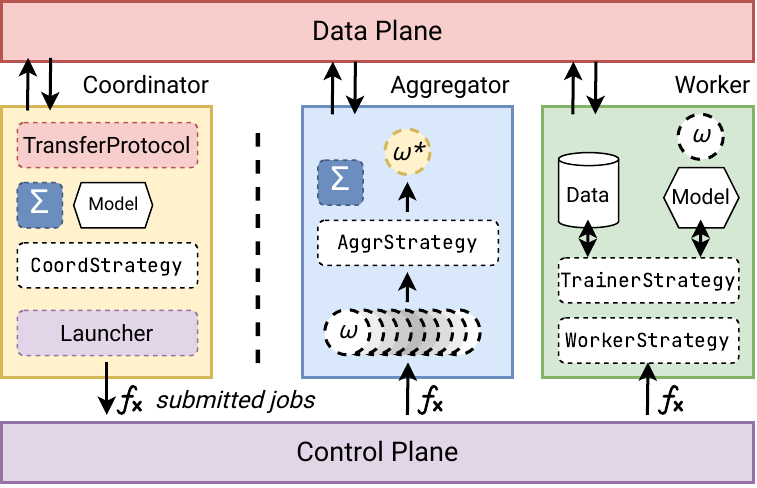}
    \caption{
        High-level view of \flox{} architecture. The \coordinator{} launches jobs to be run on \aggrs{} and \workers{} through the control plane, while data (e.g., model parameters, $\params$) are transferred through a data plane. Each \worker{} trains its local copy of the model and sends back its locally-updated model to its parent (either the \coordinator{} or an \aggr{}). Each \aggr{} aggregates the responses of its children (\workers{} and other \aggrs{} alike). The \coordinator{} facilitates the entire process.
        % \nathaniel{This is not \textit{finished} yet, still need to polish (e.g., font sizes). \textbf{Please} leave opinions so I can make changes now rather than later.} \kyle{Would be good to design as a 2 col figure. I think needs a clearer distinction between the coordinator and the agg/workers}
    }
    \label{fig:system}
\end{figure}

% Following the discussion from \autoref{sec:existing_frameworks}, 
% There are several features that are important to provide in an FL framework:
% (i) modularity to support design of custom FL algorithms and logic;
% (ii) support for common ML models and datasets;
% (iii) native support for simulating FL processes for rapid prototyping and ideation;
% (iiii) the ability to launch FL processes on real-world devices.
\flox{} is an open-source Python library for implementing HFL processes.
It is designed to be robust, scalable, and flexible %, in terms of, for example,
with respect to deep learning models, how aggregation and training tasks are launched,
%(for simulation and production), 
and how parameters are transferred between devices. 
A high-level overview of \flox{}'s design can be found in \autoref{fig:system}.
%, and how models are aggregated. 

\subsection{\flox{} Network Topologies}
\label{sec:node_roles}

% \kyle{I think we need to make it clear that these things are implemented as python objects and forward ref}
Here, we introduce the programmatic abstractions for defining networks of connected devices in \flox{}. 
The network topology is defined as a directed graph using NetworkX~(NX)~\citep{networkx}.
% Following the earlier discussion from \autoref{sec:back}, 
We consider three \textit{entity} types in \flox{} networks: 
% 
% \begin{enumerate*}[label=\textit{(\roman*)}]
    \textit{(i)} \coordinator{},
    \textit{(ii)} \aggr{},
    and
    \textit{(iii)} \worker{}.
% \end{enumerate*}
% 
The type for each entity is assigned to the nodes in the underlying NX graph as an \texttt{enum} attribute.

\begin{figure*}[t]
    \includegraphics[width=0.95\linewidth]{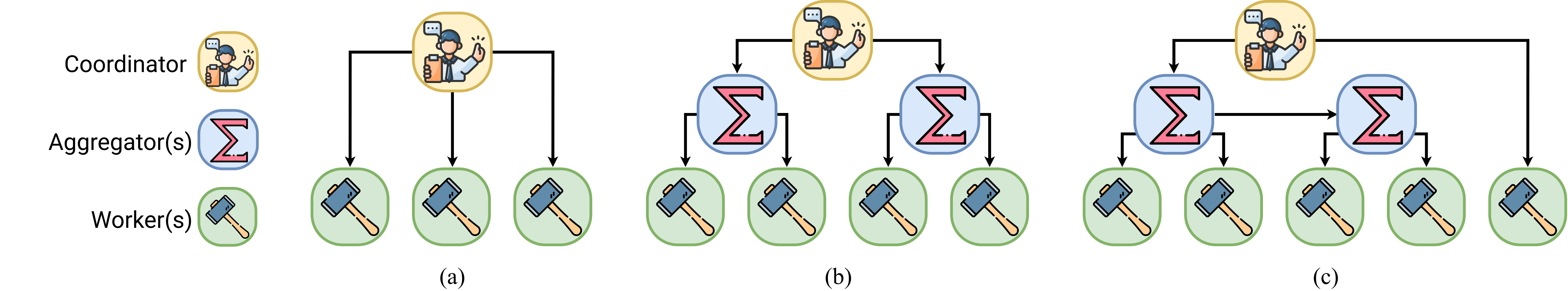}
    
    \caption{
        Example legal \flox{} network topologies: (a) simple two-tier network; (b) simple three-tier hierarchical network; (c) complex hierarchical network.
    }
    \label{fig:topologies}
\end{figure*}

% \subsubsection{Roles of Each Node}
% \label{sec:node_roles}
Each entity type is implemented as a Python class with an extensible API enabling users to customize behavior. Each entity type is responsible for different tasks related to the execution of an HFL process. 
At a high level, the \coordinator{} has three key responsibilities: 
% 
% \begin{enumerate*}[label=\textit{(\roman*)}]
    \textit{(i)} maintaining the global model;
    \textit{(ii)} submitting the appropriate training and aggregation jobs (implemented as pure functions) to the \aggrs{} and \workers{};
    and
    \textit{(iii)} acting as the \textit{global aggregator} to aggregate the model parameters returned from each direct child (\aggrs{} or \workers{}).
% \end{enumerate*}
% 
Both the \aggr{} and \workers{} are entities that wait to receive jobs to run from the \coordinator{}.
\workers{} simply train a local copy of the global model on local data. 
%the data locally-owned by that \worker{}.
\workers{} then return a \result{}---a data class that contains the locally-updated model and other information related to the completion of the job (e.g., loss, time). The \result{} is sent to its parent node (either an \aggr{} or the \coordinator{}).
\aggr{} jobs instantiate a \future{} with a callback that blocks until the \futures{} for the jobs of the \aggr{}'s children have completed. \aggr{} then will also return a \result{} to its parent.
Using a common \result{} for \aggrs{} and \worker{}s allows for \aggrs{} to aggregate what is returned to them regardless of whether they are returned by a \worker{} or another \aggr{}. 
This design choice generalizes to support arbitrary hierarchical scenarios.

\begin{lstlisting}[
    caption={
        \flox{} network topology definition as a \texttt{yaml} file. 
        Each node is defined as a dictionary with its key as its ID and various attributes: \texttt{kind}, \texttt{children},         \texttt{globus\_compute\_endpoint}, and \texttt{proxystore\_endpoint}. 
        \\
    }, 
    linewidth=0.985\columnwidth,
    label={lst:topo_yaml}, 
    float=t,
    boxpos=c,
]
MyCoordinator:
  kind: coordinator
  children: [ Aggr, Worker3 ]
  globus_compute_endpoint: null
  proxystore_endpoint: <UUID>
Aggr:
  kind: aggregator
  children: [ Worker1, Worker2 ]
  globus_compute_endpoint: <UUID>
  proxystore_endpoint: <UUID>
Worker1:
  kind: worker
  children: []
  globus_compute_endpoint: <UUID>
  proxystore_endpoint: <UUID>
...
\end{lstlisting}

%%% NOTE: Original `fedavg.tex` code snippet place

\begin{comment}
The \aggr{} nodes are \enquote{middle aggregators} that aggregate the returned model parameters of their children nodes (either \workers{} or others \aggrs{}). The \workers{} are the nodes that train their copy of the global model using their local, private data. 
More information about the jobs run on \aggr{} and \worker{} nodes is presented in \autoref{sec:launcher}.
 This is discussed earlier in \autoref{sec:back}. Succinctly, the client node is responsible for overseeing the execution of the entire HFL process. It is responsible for initializing the global model, setting up connections to the remote nodes (both aggregators and workers), and submitting jobs to be run on the other nodes in the network. The worker nodes are responsible for training the DNN being trained with their local data. The aggregator nodes are responsible for aggregating the model parameters returned by their children. 
In \flox{}, we define a data class of \texttt{JobResult} which is used to standardize what is returned by both aggregator and worker nodes. 
Because both aggregator and worker nodes return the same object, we are able to enable hierarchical execution seamlessly and with few constraints on the topology.
Aggregators will be tasked with aggregating the parameters included in the \texttt{JobResult} objects returned by their children, regardless of if those children are workers or other aggregators.
\end{comment}

% \subsection{Defining and Configuring \flox{} Network Topologies}
We define a legal network topology as follows. 
1) The network must be a rooted directed tree. 
2) There must be exactly one \coordinator{} node and it must be the root of the tree. 
3) There must be at least one \worker{} node and each \worker{} node must be a leaf of the tree. 
4) An \aggr{} node can be neither the root nor a leaf of the tree; \aggrs{} are also not required. 
% So long as the topology of a \flox{} network follows these rules listed above, it will be treated as a legal network topology and can be used for FL processes. 
These rules enable many different types of hierarchical topologies (see \autoref{fig:topologies}). 
% On top of these topological rules, 

\flox{} topologies are defined using \texttt{yaml} files. An example of such a definition is shown in \autoref{lst:topo_yaml}. 
Entities in a \flox{} topology have three properties: 
% 
% \begin{enumerate*}[label=\textit{(\roman*)}]
    \textit{(i)} an identifier, 
    \textit{(ii)} a flag indicating the entity type (i.e., coordinator, aggregator, or worker), 
    and
    \textit{(iii)} a list of children's identifiers.
% \end{enumerate*}
% 
Optionally, if deploying on remote resources, the entity may include a Globus Compute endpoint ID and a ProxyStore endpoint ID for managing remote computation and data, respectively. These properties are discussed  
% The latter two properties are related to the job launching and data transfer discussed later 
in \autoref{sec:remote} and \autoref{sec:proxystore}, respectively.  

Topology files can be loaded and used in \flox{} by \texttt{flight.Topo.from\_yaml({<filename>})}.
Using NetworkX~\citep{networkx} to define system topologies gives \flox{} the ability to load (or generate) a large variety of topology definitions.
%So long as they adhere to the topology rules, they can be used to launch \flox{} simulations. If they have the the endpoint IDs for Globus Compute and ProxyStore, then they can be quickly used to launch on remote devices.
% \nathaniel{Expand more on this, also mention NX's generative functions, and the other utility functions we provide.}

\subsection{Control Flow: Job Launching}
\label{sec:launcher}

As mentioned earlier, in \flox{}, the \coordinator{} is responsible for managing the execution of the HFL process. 
% It is worth noting that \flox{} applies a \textit{coordinator-driven} approach to launching jobs on target resources. 
% The \coordinator{} is responsible for
This includes the task of telling \workers{}  to locally train a model and \aggrs{} to aggregate models. Other FL frameworks often apply a \worker{}-driven approach, as discussed in \autoref{sec:existing_frameworks}. 
That is, it is responsible for launching the training and aggregation tasks on the target entities (e.g., a local thread or a process on a remote device). 
In \flox{} this is handled by a \launcher{}---a Python object that implements \flox{}'s \launcher{} interface for running jobs on arbitrary computing resources. 
The \launcher{} %interface 
is an asynchronous interface to submit a Python function and any input arguments for execution on a target entity (we call the executing training or aggregation function a \textit{task}). The interface returns a \textit{future} to the \coordinator{}, enabling it to monitor execution and receive a callback when the task completes. \textit{Futures} are also passed from the \coordinator{} to \aggrs{} such that they too can wait on a callback from a \worker{} or other \aggr{}. 
The result wrapped in the \textit{future} includes returned objects or exceptions.
%, from local processes to remote edge devices and HPC systems. 
%to endpoints to be run. 
We implement the \launcher{} interface on top of Python's \textit{concurrent.futures} \texttt{Executor} interface with additional requirements on the \texttt{submit()} method (e.g., submitting a \job{} and returning a \result{}). 
% The \launcher{} interface requires only the implementation of a \texttt{submit()} method. 
\flox{} supports several modes for launching jobs necessary for an FL process. 
For convenience, \flox{} includes \launcher{} implementations for three use cases:
% 
% \begin{enumerate*}[label=\textit{(\roman*)}]
    \textit{(i)} local single-node simulation via threads or processes, 
    \textit{(ii)} local multi-node simulation via Parsl~\citep{parsl},
    and
    \textit{(iii)} remote execution via Globus Compute~\citep{funcx}.
% \end{enumerate*}
% 
For convenience, \flox{} provides a high-level function for launching any type of FL process.
A small example of launching an FL process in \flox{} is shown in \autoref{lst:quickstart}.
%For instance, in APPFL~\citep{appfl} and Flower~\citep{flower}, the end devices (i.e., \workers{}) are responsible for contacting the server and establishing a connection to which the server then incorporates them into the FL process.

\subsubsection{Single-node simulation}
\flox{} provides a \texttt{LocalLauncher} that implements the \texttt{Launcher} interface using Python's standard \textit{concurrent.futures} \texttt{ProcessPoolExecutor} and \texttt{ThreadPoolExecutor}. 
%which come included in Python. 
The \texttt{LocalLauncher} enables rapid prototyping for algorithm design, experimentation, simulations, and debugging tasks related to FL/HFL.

\subsubsection{Multi-node simulation}
\flox{} implements a \texttt{ParslLauncher} to run FL/HFL processes on high-performance computing systems. Parsl~\citep{parsl}, a scalable parallel programming library for Python, is designed to run workloads on parallel and distributed computing systems (e.g., institutional clusters, supercomputers, and clouds). 
Parsl's modular architecture defines an extensible \texttt{Executor} interface via which different runtime executors can be used for different scenarios: e.g., HighThroughputExecutor (HTEX)~\citep{parsl}, RADICAL-Pilot~\citep{merzky22radical}, and WorkQueue/TaskVine~\citep{slydelgado23taskvine}. These executors implement the same asynchronous API but differ in how tasks are executed. 
%For example, HTEX implements a Python-based pilot job model in which Python worker processes are deployed on provisioned compute resources. 
Parsl supports provisioning of resources from various compute resources, including batch schedulers (e.g., Slurm, PBS), clouds (e.g., AWS), and container orchestration systems (e.g., Kubernetes).
\flox{}'s \texttt{ParslLauncher} instantiates a Parsl process using a user-defined Parsl configuration (e.g., specifying executor options such as batch queue, account, walltime). \flox{} tasks are then submitted to Parsl for execution and results are retrieved via the returned future.

%Argonne Leadership Computing Facility). 
%\nathaniel{Yadu}

\begin{lstlisting}[
    caption={
        Example \flox{} program that showcases a simple \flox{} program.
        \autoref{lst:topo_yaml} provides an example \texttt{my-topo.yaml} file.
    }, 
    language=Python,
    label={lst:quickstart}, 
    float=t,
    boxpos=c,
    linewidth=0.985\columnwidth,
]
import torch
import flight as fl

from torchvision.datasets import MNIST

class MyModule(fl.TorchModule):
  # Model definition.

topo = fl.Topo.from_yaml('my-topo.yaml')
mnist = MNIST(...)
fed_data = fl.federated_split(
  topo, mnist, num_classes=10,
  sample_alpha=1.0, label_alpha=1.0
)
trained_model, results = fl.federated_fit(
  topo, 
  MyModule(), 
  fed_data, 
  strategy='fedavg', 
  where='local', # 'globus-compute',
  kind='sync',   # 'async'
)
results.to_csv('my-results.csv')
\end{lstlisting}

\subsubsection{Execution on remote endpoints}
\label{sec:remote}

\flox{} includes a \texttt{GlobusComputeLauncher} to enable simple and convenient remote execution of functions for FL/HFL processes. 
% \flox{} use Globus Compute~\citep{funcx} for our FaaS implementation as it provides a unique hybrid and federated deployment model.
Globus Compute~\citep{funcx} implements the Function-as-a-Service (FaaS) paradigm enabling execution of Python functions. It combines a single cloud-hosted service with an ecosystem of user-deployed \textit{endpoints}. Thus, \flox{} can use the cloud-hosted Globus Compute service to orchestrate execution of tasks (e.g., \aggr{} and \worker{} tasks) on arbitrary remote devices. As the endpoint software is lightweight---a pip-installable Python agent that communicates via cloud-hosted message queues---it is easily deployed on diverse compute devices. Further, it requires only outbound connectivity to the Globus Compute cloud service, therefore addressing challenges with firewalls and Network Address Translation (NAT) used in many edge environments. 
% and is significantly less resource intensive than complete FaaS services (e.g., OpenWhisk~\citep{openwhisk}). 
The endpoint software builds on Parsl and is therefore equipped to dynamically provision and then execute tasks on diverse systems, including HPC clusters and Kubernetes. Finally, Globus Compute endpoints have been deployed nearly 10,000 times around the world~\citep{bauer2024globus}; on those systems, \flox{} can be used without needing to deploy any new infrastructure.

\flox{}'s \texttt{GlobusComputeLauncher} uses 
Globus Compute's Python SDK and executor interface to submit tasks. \flox{} users must provide OAuth~2 access tokens to instantiate the executor and they must also provide the set of compute endpoints to be used in the network topology definition. \flox{} submits \aggr{} and \worker{} tasks to Globus Compute as required and tracks results via the Globus Compute \future{} returned to \flox{}.

\subsection{Data Flow: Federated Data Transfer}
\label{sec:proxystore}

The purpose of HFL is to improve performance by distributing model aggregation to different locations. Thus, it would be both inefficient and costly if all data had to pass through the \coordinator{} rather than be passed directly between the participating entities.
% While Globus Compute provides an ideal mechanism for orchestrating the reliable execution of \worker{} and \aggr{} tasks on remote computers, there is one limitation of our architecture: the hybrid architecture requires that task input and output data be transferred via the cloud service. This would be both inefficient and costly  when considering the tree architecture of a \flox{} topology. 
% Further, Globus Compute, like most FaaS frameworks, places limits on the amount of data that can be passed to/from tasks, in this case 10MB---far lower than the size of many models. 
% It is common for severless computing frameworks to enforce data payload restrictions for the invocation of functions on remote endpoints. 
% This decision is made due to the natural data transfer costs associated with any kind of communication. 
% The payload restriction for Globus Compute is set at 5MB---which comparable to other frameworks.
% While this is a reasonable design decision, it effectively disallows the execution of most FL/HFL processes altogether. 
% This is because the size of DNNs very commonly exceed beyond the payload caps found in serverless computing frameworks. 
Our solution to this problem is to decouple the transfer of data from the execution of the task itself. We effectively use a control layer, via which small function invocations and small function results are sent via the launcher (e.g., via Globus Compute's cloud service or through Parsl's DataFlowKernel). We use ProxyStore~\citep{pauloski2023proxystore,pauloski2024proxystore} to move larger data (e.g., models and weights) directly between the tasks running on \workers{} and \aggrs{}. 
ProxyStore is %a federated data transfer service which enables the transfer of large data for serverless computing frameworks. 
a framework that uses Python proxy objects to provide
\textit{pass-by-reference} semantics in distributed computing environments. 
% This is done by creating data proxies which act as a \textit{pass-by-reference} abstraction for distributed applications. 
Given some data $x$, ProxyStore will generate a proxy object $p(x)$ for the data $x$. 
This proxy---essentially a small reference to the data---can then be sent with the computing task via the launcher. 
% This proxy is trivial in size since it essentially acts as a reference to the data.
When the data is needed by the task being executed on a device, it will then be transferred by ProxyStore using the user's choice of transfer protocol. ProxyStore supports many data transfer protocols, including Redis and \textit{Remote Direct Memory Access}~(RDMA). 
It also implements a peer-to-peer transfer solution, referred to as the \texttt{EndpointConnector}, for direct communication between endpoints. This transfer mechanism uses UDP hole punching to establish connections between devices that are behind firewalls and that use NAT for private networks. 
% \flox{} takes advantage of this backend for supporting peer-to-peer communication of data (i.e., model parameters) across hierarchical networks. 
% Though, it is worth noting that other supported backends can be integrated for future works.
% \kyle{We could consider a picture here to show the control vs data side}

\subsection{Execution Schemes: Synchronous and Asynchronous FL}
Now, we describe \flox{}'s execution schemes to perform FL/HFL either synchronously or asynchronously. Specifically, we describe how jobs are launched and the timing of when their returned results are aggregated. It is worth noting that, for both execution schemes, the role of the \coordinator{} remains the same. Its role is to oversee and manage the execution during the lifetime of the FL/HFL process.

\flox{}'s modular design allows for customization for different FL algorithms and strategies. 
This modularity is achieved via the \texttt{Strategy} abstraction. As discussed more fully in \autoref{sec:strat},
a \texttt{Strategy} provides callbacks that can be programmed by users to customize the execution of their FL/HFL processes. 
For the sake of brevity, we forego mentioning \textit{all} available callbacks when describing the execution scheme.

\subsubsection{Synchronous HFL~(SHFL)}
In the SHFL case, \flox{} begins a series of \enquote{rounds}.
In each round, the \coordinator{} selects \workers{} to participate in the round (i.e., to train their local model). % global model. 
Depending on the configuration, the selected workers can be either a subset of, or all, workers.
The \coordinator{} then identifies all \aggrs{} that are on the path from the \coordinator{} to each selected \worker{}.
If all \workers{} are selected to perform local training, then it follows that all \aggrs{} are also selected.
Next, the \coordinator{} submits jobs to the selected \workers{} and the relevant \aggrs{} by using its configured \launcher{}.
These submissions are generated in a breadth-first search-like fashion where the \coordinator{} (at the root) traverses each of its children and submits the appropriate (training or aggregation) job to each, retaining a corresponding \texttt{Future} for each child. 
The \coordinator{} submits aggregation jobs to each \aggr{},
%as well and submits their aggregation job 
including for each a list of the \aggr{}'s children's \texttt{Futures} as an argument. The \aggrs{} begin their aggregation jobs once their children's respective \texttt{Futures} have completed. Because we separate the data flow from the control flow via ProxyStore (see \autoref{sec:proxystore}), the data from the results of the child \texttt{Futures} for each \aggr{} is never sent back to the \coordinator{}. The data (i.e., model parameters) are transferred to the \aggr{} that depends on it.
When \workers{} are traversed as the leaves of the tree, the \coordinator{} submits a local training job to these entities. Each of which will eventually return a \texttt{JobResult} to its parent entity.
The \coordinator{}, like the \aggrs{}, waits for the completion of its children's \texttt{Futures}. 
When the \texttt{Futures} complete, the \coordinator{} aggregates to update the global model.
It then performs additional tasks (e.g., evaluating the global model, processing results) and restarts the process until a set number of rounds concludes.

\subsubsection{Asynchronous FL~(AFL)}
\flox{} currently supports only two-tier AFL topologies. Supporting AFL with arbitrary hierarchies would require implementing a new type of launcher with the ability to maintain stateful entities at each level.
In SHFL, the \coordinator{} and \aggrs{} remain idle until all their children's \texttt{Futures} resolve. 
% AFL does not take on this approach. 
% In \flox{}, the AFL implementation has the 
Specifically, the \coordinator{} launches local training jobs on each \worker{} and maintains a list of \texttt{Futures}. 
The \coordinator{} takes action as each \future{} is completed. 
As soon as one \future{} completes, the \coordinator{} retrieves the locally-updated model parameters from the \future{} result and performs a partial aggregation to update the global model---see \autoref{eq:async_avg}.
The \coordinator{} then launches a new local training job on the worker whose \future{} just completed and appends its new \future{} to its list of \texttt{Futures}.

\subsection{Enabling Machine Learning}

We describe now how \flox{} works with existing deep learning frameworks to implement, train, and use models and manage datasets.

\textbf{Deep learning models.}
\flox{} relies on the PyTorch~\citep{pytorch} framework to define, train, and evaluate AI models. We chose PyTorch due to its ubiquity in both deep learning and FL. % as well as for applied ML use cases. 
\flox{} implements the \texttt{\flox{}Module} class which extends the \texttt{torch.nn.Module} class for defining a neural network. This class offers callbacks similar to the \texttt{LightningModule} from PyTorch Lightning~\citep{pytorch_lightning}. Further, the \texttt{\flox{}Module} requires only that two of these callbacks are implemented by the user (i.e., \texttt{training\_step()} and \texttt{configure\_optimizers()}).  \flox{} can then train the model by launching remote jobs on \workers{} without further specification from users. We choose to implement the \texttt{\flox{}Module} class rather than use Lightning's default \texttt{LightningModule} so as not to require the latter as a dependency given its size and complexity---an important consideration when deploying on edge devices. 
However, in future work we will add 
support for PyTorch Lightning.
%as an alternative backend for local training of models. 
% Additionally, if users have specific details for how to train their models with \flox{}, they can write their own function to be sent to \workers{}. 
\flox{} also provides a callable \texttt{LocalTrainJob} class that provides a common interface to define how models are trained on \workers{} to allow highly-custom local training loops. 
% To this end, users are free to write their own custom pure functions that implement this interface's function signature and then give that function to \flox{} as an input. 
% It will then use that in lieu of its default implementation.

\textbf{Datasets} in \flox{} are based on the PyTorch \texttt{Dataset} abstraction. \flox{} requires that each \worker{} have its own data. For simulated FL processes, this is typically based on partitioning a common dataset (e.g., FashionMNIST~\citep{fashion_mnist})  into separate subsets for each worker{}. \flox{} simplifies this process by providing a class (\texttt{FederatedSubsets}) that allows users to define how a dataset is partitioned and allows users to easily control the distribution of data across \workers{}.
%partitioning a dataset into what we call a  
This class is effectively a dictionary where the key is the \worker{} ID and the value is a PyTorch \texttt{Subset} of the data. 
Crucial to the needs of HFL deployments, \flox{} also allows users to load data local on the \workers{}.

\section{Strategies}
\label{sec:strat}

FL is an active research domain and new algorithms are frequently proposed to meet
new requirements relating to system heterogeneity, data/statistical heterogeneity, and many other challenges. 
FL frameworks must thus be flexible and customizable to meet the needs of both FL researchers and practitioners so that novel and custom algorithms can be readily implemented and deployed.
\flox{} implements a modular and extensible abstraction for specifying such algorithms, which we refer to as a \strategy{}.
A \strategy{} is essentially a wrapper for an FL algorithm or solution that provides specific implementation details necessary to execute the HFL process. In designing this architecture we considered the needs of various aggregation methods, asynchronous and hierarchical FL, and privacy-preserving FL. 
%The definition of a \strategy{} class is presented in \autoref{lst:strategy}. 

\begin{lstlisting}[
    caption={Definition of the \texttt{Strategy} class in \flox{}.}, 
    language=Python,
    label={lst:strategy}, 
    float=t,
    linewidth=0.985\columnwidth,
    boxpos=c,
]
@dataclass(frozen=True)
class Strategy:
    coord_strategy  : CoordinatorStrategy
    aggr_strategy   : AggregatorStrategy
    worker_strategy : WorkerStrategy
    trainer_strategy: TrainerStrategy
\end{lstlisting}

% \subsection{Strategy Components}
% \label{sec:strategy_comps}
% \ian{I found the following a bit convoluted and repetitive. Not sure how to change though.}

% {\color{blue}

Specifically, a \strategy{} is made up of the following components:
    \textit{(i)} \coordstrat{}, 
    \textit{(ii)} \aggrstrat{}, 
    \textit{(iii)} \workerstrat{}, 
    and 
    \textit{(iv)} \trainerstrat{} (see \autoref{lst:strategy}).
We refer to these components as \enquote{sub-strategies} because they implement the respective logic that is run on all of the entities in a topology for an HFL process.
The \coordstrat{} is run on the \coordinator{} and provides callbacks for users to implement custom worker selection algorithms based on worker conditions.
The \aggrstrat{} is run on the \aggrs{} and provides callbacks for custom parameter aggregation algorithms.
The \workerstrat{} and \trainerstrat{} sub-strategies are both run on the \workers{}; the former provides callbacks more specific to the execution of the entire local training job performed by the worker (e.g., caching/recording system conditions, pre-processing data on the worker) while the latter provides callbacks specific exclusively to the training loop (e.g., modifying the loss before back-propagation).

A new \flox{} strategy can be defined simply by composing four existing sub-strategies.
Or, a user can provide custom implementations of one or more of the sub-strategy classes.
% discussed in \autoref{sec:strategy_comps}.
To simplify the development of custom implementations, \flox{} provides a default implementation for each sub-strategy; a user can inherit from such a default implementation and override only what is needed.
As an example, we present our  \texttt{FedAvg}~\citep{fl_seminal_1} implementation in \autoref{lst:fedavg}.
For user convenience, \flox{} provides implementations of 
strategies for common FL algorithms, including \texttt{FedSGD}, \texttt{FedAvg}~\citep{fl_seminal_1}, \texttt{FedProx}~\citep{fedprox}, 
% \texttt{AgnosticFL}~\citep{agnostic_fl}, 
% \texttt{FedAdam}~\citep{fedadam}, 
and \texttt{FedAsync}~\citep{fedasync}. % implemented as strategies.

\begin{lstlisting}[
    caption={
        This custom implementation of the \texttt{FedAvg} algorithm is constructed by subclassing the default implementations of \texttt{AggregatorStrategy} and \texttt{WorkerStrategy}. The required callbacks are then overriden and those implementations are wrapped in the \texttt{FedAvg} class which contains all the logical components. Note the use of the \texttt{FedSGD} algorithm's \texttt{CoordinatorStrategy} on line~24.
    }, 
    language=Python,
    label={lst:fedavg},
    linewidth=0.985\columnwidth,
    float=t,
    boxpos=c,
]
class FedAvgAggr(DefaultAggregatorStrategy):
  def aggregate_params(
    self, state, children_states, children_params
  ) -> Params:
    weights = {}
    for node, child_state in children_states.items():
      weights[node] = child_state['num_data_samples']
    state['num_data_samples'] = sum(weights.values())
    return average_params(  
      # Flight-provided utility fn
      children_params, weights=weights
    )

class FedAvgWorker(DefaultWorkerStrategy):
  def before_training(
    self, state: WorkerState, data
  ) -> tuple[WorkerState, t.Any]:
    state['num_data_samples'] = len(data)
    return state, data

class FedAvg(Strategy):
  def __init__(self, **kwargs):
    super().__init__(
      coord_strategy=FedSGDCoordinator(**kwargs),
      aggr_strategy=FedAvgAggr(),
      worker_strategy=FedAvgWorker(),
      trainer_strategy=DefaultTrainerStrategy(),
    )
\end{lstlisting}

% List out the available callbacks.
\subsection{Defining Custom Strategies}
\label{sec:custom_strategies}

\section{Evaluation}
\label{sec:eval}

% \nathaniel{Preface?}
% \input{figures/weak_scaling_flox}
% \input{figures/tasks_per_sec}

%%%%%%%%%%%%%%%%%%%%%%%%%%%%%%%%%%%%%%%%%%%%%%%%%%%%%%%%%%%%%%%%
%%%%%%%%%%%%%%%%%%%%%%%%%%%%%%%%%%%%%%%%%%%%%%%%%%%%%%%%%%%%%%%%
%%%%%%%%%%%%%%%%%%%%%%%%%%%%%%%%%%%%%%%%%%%%%%%%%%%%%%%%%%%%%%%%

\begin{figure*}[ht]
    \centering
    \includegraphics[width=\linewidth]{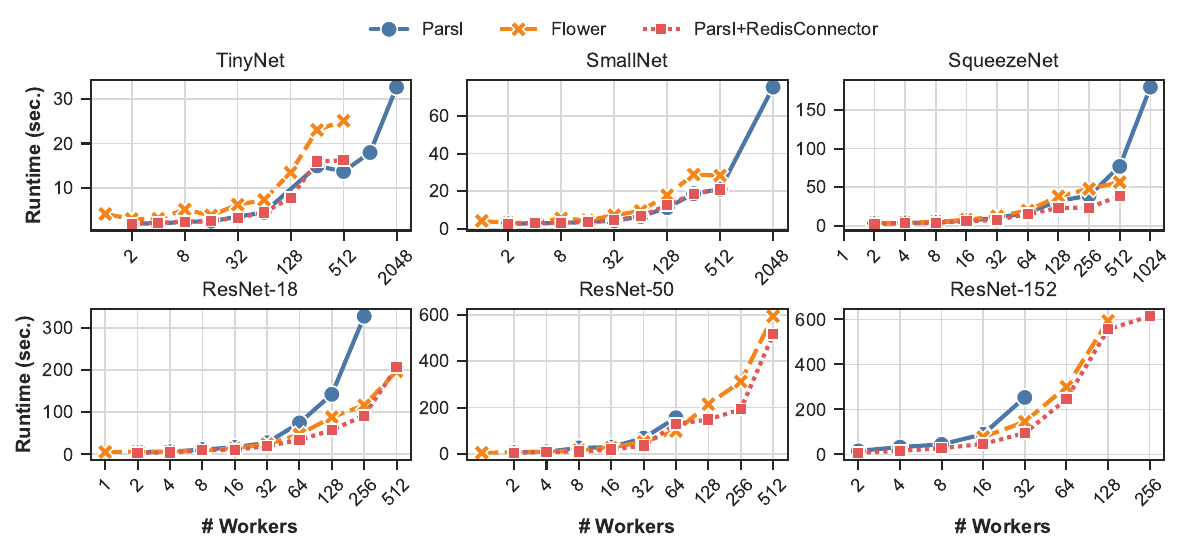}
    \caption{
        Weak scaling results: Runtime of Flower vs. \flox{} using Parsl and Parsl+RedisConnector for a series of increasingly complex models (see \autoref{table:models}).
        Results confirm that our proposed \flox{} framework provides better performance and, in some cases, also scales to more workers. % running in parallel.
    }
    \label{fig:weak_scaling}
\end{figure*}
\begin{table}[t]
\small
\centering
\begin{tabular}{lcc}
    \toprule
    \textbf{Model} & \textbf{Params} & \textbf{Size}  \\
    \midrule
    TinyNet             
        & 2 & 8 bytes \\
    SmallNet~\citep{smallnet}
        & 62K & 242 KB \\
    SqueezeNet~\citep{iandola2016squeezenet}
        & 1.2M & 5 MB \\
    ResNet-18~\citep{resnet}
        & 11M & 45 MB \\
    ResNet-50~\citep{resnet}
        & 23M & 98 MB \\
    ResNet-152~\citep{resnet} 
        & 60M & 231 MB \\
    \bottomrule
\end{tabular}
\caption{Sizes of the models considered in our experiments.}
\label{table:models}
% \vspace{-2em}
\end{table}

% \begin{table}[t]
% \centering
% \scriptsize
% \caption{Sizes of the models considered in our experiments.}
% \label{table:models}
% \begin{tabular}{rllllll}
% \toprule
%        & TinyNet & SmallNet & SqueezeNet & ResNet-18 & ResNet-50 & ResNet-152 \\
% \midrule
% Params & 2       & 62K      & 1.2M       & 11M       & 23M       & 60M        \\
% Size   & 8 bytes & 242 KB   & 5 MB       & 45 MB     & 98 MB     & 231 MB \\
% \bottomrule
% \end{tabular}
% \end{table}

\begin{figure}[h]
    \centering
    \includegraphics[width=0.9\linewidth]{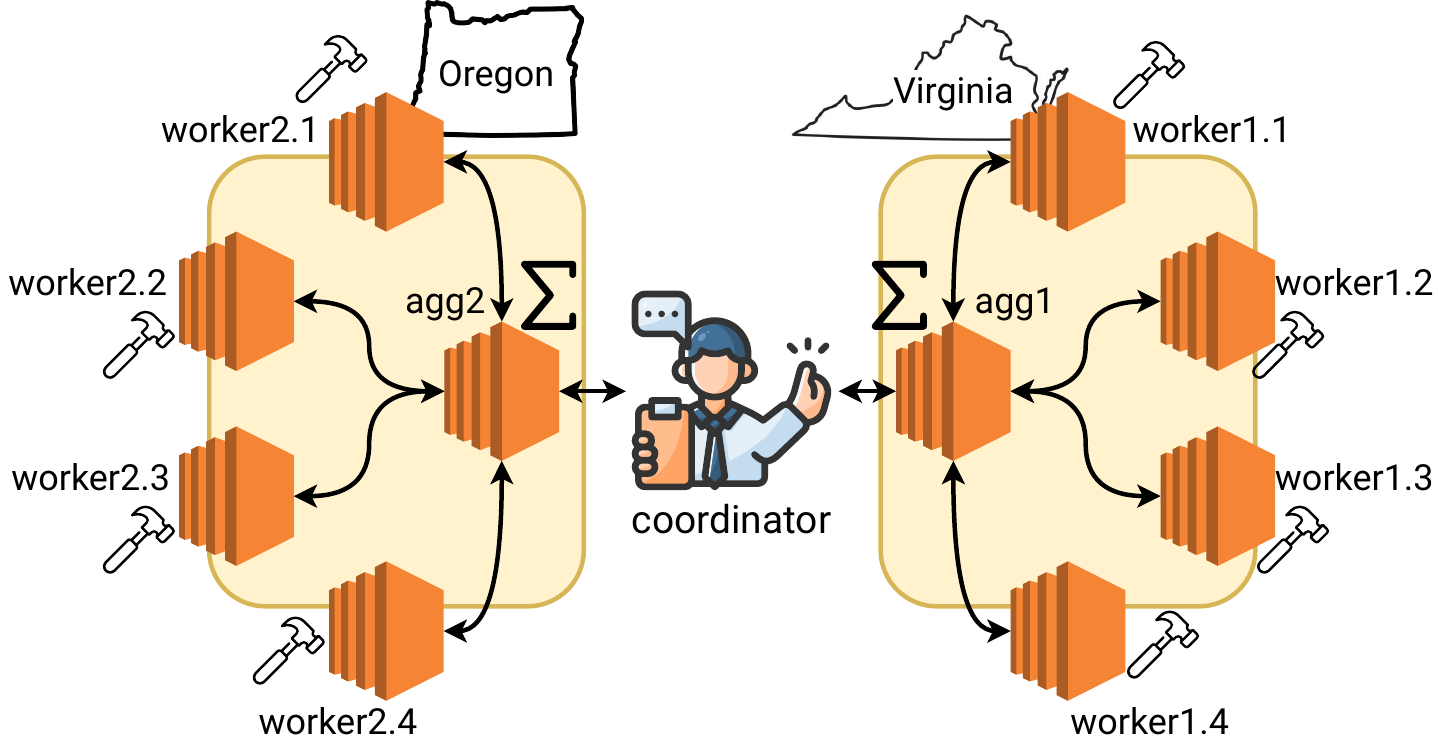}
    \caption{
        The hierarchical topology of the EC2 testbed employed for remote tests with \texttt{GlobusComputeLauncher} and ProxyStore.
    }
    \label{fig:ec2_topo}
\end{figure}

We evaluate \flox{}'s scalability in multi-node experiments on an HPC cluster, hierarchical communication costs and asynchronous FL makespan in single node experiments, and performance in on-device experiments conducted on AWS. 

\subsection{\flox{} Scaling Tests}
\label{sec:scaling_test}
% In \autoref{sec:launcher}, we mention the requirement for \flox{} to support high-throughput simulations to support rapid research prototyping and ideation. To this end, 
We explore how \flox{} scales with different model sizes and compare with the state-of-the-art Flower framework. 
% We perform a weak scaling test using increasing sizes of deep neural networks with both our proposed \flox{} framework and the well-established Flower framework.
%Because FL is naturally constrained by the number of workers in the network, it does not lend itself well to strong scaling tests.
For a fair comparison against Flower, we consider only two-tier topologies and scale the number of workers and model size. 
%where we increase the number of workers in the simulated topology.

%\noindent
\textbf{Testbed:} 
We use SDSC Expanse, a 5.16 petaflop cluster with 728 CPU compute nodes, each with two 64-core AMD EPYC 7742 processors and 256~GB memory. The nodes are connected with a 100~Gb/s InfiniBand interconnect.

%\noindent
\textbf{Models:} 
We use the six image classification model architectures in \autoref{table:models} to evaluate \flox{} performance and overheads. 
TinyNet, the smallest model, comprises a single linear layer with 1 input channel and 1 output channel. 
%to evaluate worst case performance. 
SmallNet is a small neural network from the Pytorch documentation~\citep{smallnet}. 
The ResNet~\citep{resnet} and SqueezeNet~\citep{iandola2016squeezenet} implementations are those included in PyTorch.

%\noindent
\textbf{Configuration:} 
We configure \flox{} and Flower with a two-tier hierarchy by deploying a \coordinator{} and server, respectively, on a single compute node. We then deploy workers on separate compute nodes and increase the number of workers from 1 to 128 (each using a single core on a single node) and then up to 2048 (on 16 nodes). To evaluate scaling performance and overhead we reduce the fixed processing costs to measure worst-case performance. To do so, we configure both frameworks such that the clients do not train or evaluate the model and instead simply pass a model with randomized weights back to/from the coordinator. We use \texttt{FedAvg} to aggregate the models. 
We record the time from when the clients start until the aggregator has received and aggregated all models.

In \flox{}, we configure the \coordinator{} with a single model with randomized weights. \flox{} then distributes that model to each \worker{} who instantiate and then return the model. We configure \flox{} with the \texttt{ParslLauncher} and compare Parsl with and without ProxyStore to measure the benefits of decoupling control from data plane. We configure ProxyStore to use the RedisConnector and deploy Redis on the same node as  the \coordinator{}. Flower is a client-driven framework and thus we first initialize random models at each client. The server then picks a client at random and distributes their model weights to the other clients. 
%Thus, Flower has one additional single network communication of the model than \flox{}.

% We run experiments only once due to the cost to run repeated experiments.

\textbf{Results:}
\autoref{fig:weak_scaling} shows weak scaling results as we increase the number of workers involved in training. We see that \flox{} with Parsl outperforms Flower for smaller models, achieving faster results and scaling beyond Flower (2048 compared to 512 workers). As we scaled Flower to 1024 and 2048 workers we observed  gRPC errors that prevented aggregation. We did not scale Parsl beyond 2048 nodes due to limited allocation. For the larger ResNet models, Flower slightly outperformed \flox{} with Parsl, a difference that we attribute to the efficiency of the gRPC protocol and the fact that Parsl is not optimized to move large data volumes over its ZMQ-based protocol. \flox{} with ProxyStore's RedisConnector overcomes this limitation and we see better runtime and scalability than Flower for all models considered.

%%%%%%%%%%%%%%%%%%%%%%%%%%%%%%%%%%%%%%%%%%%%%%%%%%%%%%%%%%%%%%%%
%%%%%%%%%%%%%%%%%%%%%%%%%%%%%%%%%%%%%%%%%%%%%%%%%%%%%%%%%%%%%%%%
%%%%%%%%%%%%%%%%%%%%%%%%%%%%%%%%%%%%%%%%%%%%%%%%%%%%%%%%%%%%%%%%

\subsection{Distributed On-Device Deployment}
\label{sec:remote_eval}

\begin{figure*}[ht!]
    \centering
    \includegraphics[width=\linewidth]{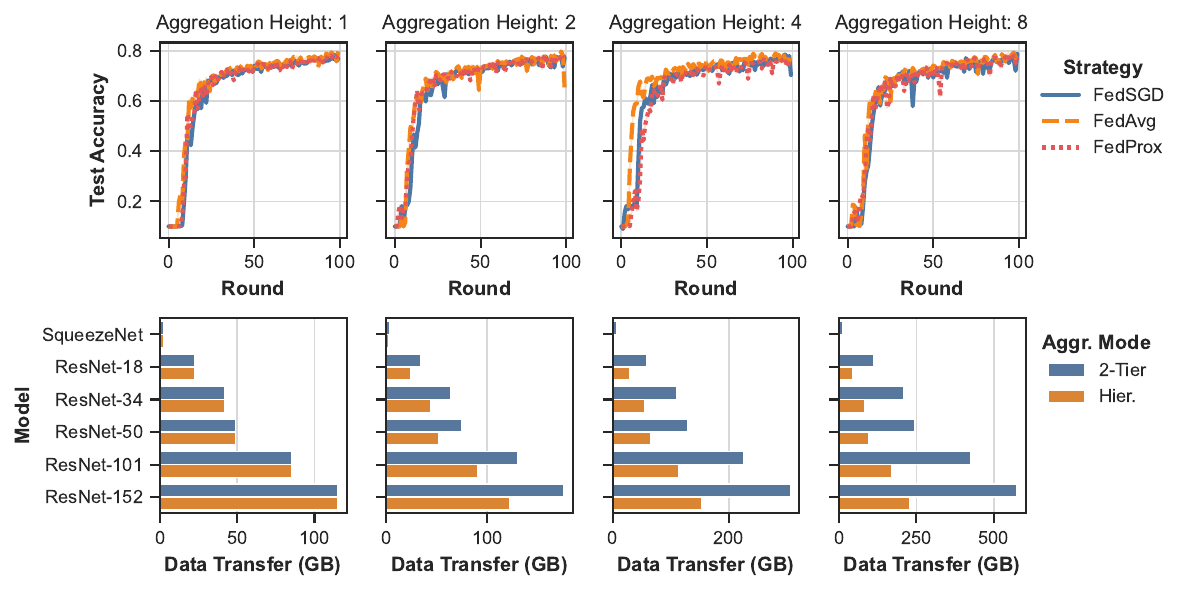}
    \caption{
        \emph{Top row}:  
            Test accuracy while training over the FashionMNIST dataset using three different strategies. Testing is done on the \coordinator{}.
        \emph{Bottom row}:
            Comparison of the total data transferred across the network (implemented as a balanced tree with $256$ leaf workers) at each round. For both 2-tier and HFL, the data transfer analysis assumes the same communication topology, though HFL \aggrs{} aggregate the models on return to the \coordinator{}.
    }
    \label{fig:acc_comm_hier}
\end{figure*}

We now evaluate the use of \flox{} in a distributed testbed to replicate a real-world deployment. We use Globus Compute and ProxyStore. 
%for the control plane and ProxyStore for the data plane.

\textbf{Testbed:}
We constructed the three-layer, 11-node hierarchical \flox{} topology shown in \autoref{fig:ec2_topo}. 
The coordinator is deployed on an Apple Silicon M1 laptop with 32GB RAM while the eight workers and two aggregators are deployed on 10 \texttt{m2.small} EC2 instances (2vCPU, 4GB RAM) on AWS cloud. We deployed one aggregator and four workers in each of AWS's Virginia and Oregon regions.  We configured each instance with both a Globus Compute endpoint and a ProxyStore endpoint. We used ProxyStore's \texttt{EndpointConnector} for wide-area data transfer.

\begin{figure}[t]
    \centering
    \includegraphics[width=\linewidth,trim=2.5mm 2.5mm 2.5mm 2.5mm,clip]{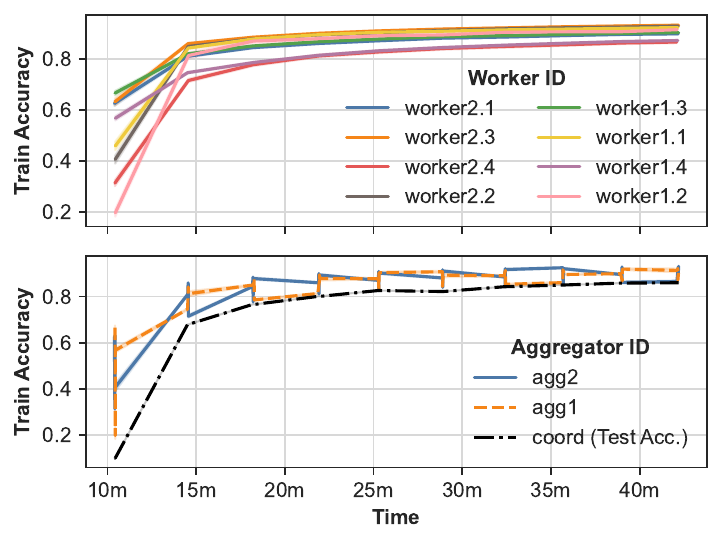}
    \caption{
        Training accuracy over time at \workers{} and \aggrs{}, and test accuracy at \coordinator{}, for an HFL application (image classification with \texttt{SmallNet}) on an 11-node distributed testbed. 
        % using \texttt{GlobusComputeLauncher} and ProxyStore in \flox{}.
    }
    \label{fig:ec2_results}
\end{figure}

\textbf{Model and Data:} 
% In this test, 
We use \texttt{SmallNet} (see \autoref{table:models}) to perform image classification on the FashionMNIST dataset.
%This model is made up of two 2D convolutional layers, a pooling layer, and three dense layers. 
For simplicity, we use the standard stochastic gradient descent algorithm for optimizing the model parameters, with a learning rate $\eta$~=~0.01.
To obtain a non-IID data distribution, as is common in decentralized data~\citep{fl_seminal_1}, we partition data across the eight \workers{} by using a dual Dirichlet distribution. 
Specifically, the number of data samples at each worker follows the Dirichlet distribution with shape parameter $\alpha$~=~3.0, and the distribution of labels across each worker follows a Dirichlet distribution with shape parameter $\alpha$~=~1.0.

\textbf{Results:}
\autoref{fig:ec2_results} shows the \textit{training accuracies} obtained over time at each worker (above) and, averaged across workers, at the two coordinators (below) when running this HFL application on our testbed.
In the lower plot, we also show the \textit{accuracy} reported by an evaluation step performed on the \coordinator{}.
The test accuracy converges with the average training accuracies of the two sets of \workers{}.

%%%%%%%%%%%%%%%%%%%%%%%%%%%%%%%%%%%%%%%%%%%%%%%%%%%%%%%%%%%%%%%%
%%%%%%%%%%%%%%%%%%%%%%%%%%%%%%%%%%%%%%%%%%%%%%%%%%%%%%%%%%%%%%%%
%%%%%%%%%%%%%%%%%%%%%%%%%%%%%%%%%%%%%%%%%%%%%%%%%%%%%%%%%%%%%%%%

\subsection{Hierarchical FL vs. Two-Tier FL: Communication Cost}
\label{sec:comm_cost_test}

HFL takes advantage of the multiple hops necessary in network topologies by using \aggrs{} to perform intermediate aggregation rather than directly traversing all links between \worker{}\texttt{s} and the \coordinator{} with all models as in two-tier systems.
We use \flox{}'s \texttt{LocalLauncher} to simulate the benefits of HFL by comparing the total amount of data transferred over the network.
%
%Naturally, transmitting data over the Internet from one device to another usually requires multiple hops even if it is abstracted from the user.
%This is also true for standard two-tier FL even if the FL process does not explicitly consider this.
%
% {\color{red}
% Originally, the math-heavy text was here.
% I pushed it to \autoref{app:tree_config}.
% TODO: Fill in the blanks to smooth over the writing.
% }
%
%In the former, each \worker{} communicates their model parameters on the network path from \worker{} to the \coordinator{}. Meanwhile, HFL has \aggrs{} aggregate all parameters returned from their children and thus communicates only one set of parameters back to its parent. 
%Thus each edge in the topology is traversed twice in HFL, once to send the models/parameters to the \workers{} and once again to return the model/parameters to the coordinator.
%
We initialize the simulation with a \text{balanced} tree topology with 256 leaves (i.e., 256 \workers{}) and vary its height from $h=1, 2, 4, 8$. We generate these tree topologies using the \texttt{balanced\_tree($\cdot$)} function from NetworkX~\citep{networkx}. 
This function takes a height~($h$) and branching factor~($b$) as an argument, which we compute by 
$b=256^{1/h}$.
% $b=\sqrt[h]{256}$.
%
Given the balanced trees, we use the model sizes from \autoref{table:models} to numerically compute the data transfer costs for two-tier FL and HFL. 
The mathematical details behind these calculations are presented in \autoref{app:tree_config}.
The reduction in transfer volume achieved by HFL is shown in \autoref{fig:acc_comm_hier}.
%As expected, the transfer volumes increase drastically with the height of the topology. 
The benefit of HFL becomes increasingly notable with topology height, saving 60.13\%  in the case of a height of 8 with the ResNet-152 model. \autoref{fig:acc_comm_hier} also shows comparable test accuracy during training across all topology heights, though test accuracy becomes less stable as height increases.

While comprehensively evaluating the impact of hierarchies on FL model performance is beyond the scope of this work, \flox{} provides a framework to enable research into such questions.

% We consider a set of decentralized network of edge devices (see \autoref{tab:gl-testbed}) with nearby aggregators and a flat version of this network with no intermediary aggregators. We then compare the communication costs associated with our proposed Hierarchical FL solution and several FL algorithms from the literature, all implemented in \flox{}.
% \todo{}

%%%%%%%%%%%%%%%%%%%%%%%%%%%%%%%%%%%%%%%%%%%%%%%%%%%%%%%%%%%%%%%%
%%%%%%%%%%%%%%%%%%%%%%%%%%%%%%%%%%%%%%%%%%%%%%%%%%%%%%%%%%%%%%%%
%%%%%%%%%%%%%%%%%%%%%%%%%%%%%%%%%%%%%%%%%%%%%%%%%%%%%%%%%%%%%%%%

\subsection{Comparing Synchronous vs. Asynchronous FL}
\label{sec:sync_vs_async}

% Asynchronous FL~(AFL) is an approach to FL where responses from worker nodes are immediately aggregated using a rolling or partial aggregation method. This means that workers are not held idle having to wait for the next round to be formally started by the coordinator. \flox{} provides support for AFL workflows out-of-the-box. Here, we showcase AFL and contrast it to synchronous FL.

We evaluate Asynchronous FL~(AFL)  on a two-tier topology with 12 \workers{} using the \texttt{LocalLauncher}. 
We partition the \mbox{FashionMNIST}~\citep{fashion_mnist} dataset across these workers with a dual Dirichlet distribution (via the \flox{}-provided \texttt{federated\_split()} utility function) such that the workers have non-IID distributions of labels and modestly imbalanced numbers of samples. 
We use the \texttt{LocalLauncher} to start 12 workers, and train SmallNet (see \autoref{table:models}) with both \texttt{FedAvg}~(synchronous) and \texttt{FedAsync}~(asynchronous). 
In both cases, we configure the FL process to run for 20 rounds (i.e., each worker locally trains the model 20 times).

\begin{figure}[t]
    \centering
    \includegraphics[width=\linewidth,trim=2mm 2mm 2mm 2mm,clip]{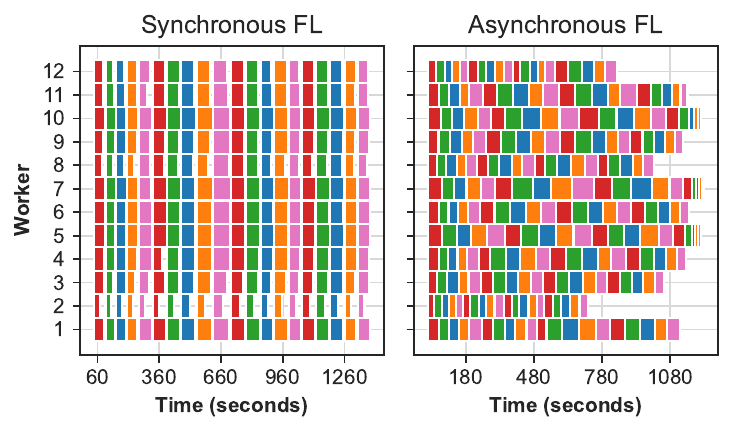}
    \caption{
    Synchronous vs.\ asynchronous FL in \flox{}. 
    Each horizontal bar indicates a worker performing local training, with color indicating round number.
    Vertically aligned white regions in the Synchronous case indicate when workers sync during aggregation before beginning the next round.
    In contrast, the Asynchronous case has clear overlaps where workers are working. 
    % \kyle{I dont see blue lines?}
    % \ian{Is it possible to remove the ".0"s in the x axes? and, put synchronous on the left, as it is discussed first?} 
     }
    \label{fig:sync_vs_async}
\end{figure}

The visualization in \autoref{fig:sync_vs_async} shows via solid colors when the workers are engaged in training in each case. %Here, we see how AFL compares to synchronous FL in \flox{}. 
We see that synchronous FL (SFL) leaves some workers idle for long periods, whereas AFL keeps the workers more active.
Overall, AFL has a nearly 20\% shorter makespan than SFL.  
Workers in the AFL run were idle, on average, just 0.061\% of the time (standard deviation 0.013\%) until they completed their final round, while workers in SFL run were idle 14.858\% of the time (standard deviation 11.313\%).
%Depending on a given use cases, AFL may prove beneficial as a result.
While a comprehensive analysis of the trade-offs between SFL and AFL is % an intricate and nuanced research question that is
beyond the scope of this paper---such analyses are the focus of related work~\citep{chen2020asynchronous, fedasync}---we find that
AFL achieved a training accuracy of 87\% whereas SFL achieved 90\% accuracy.

%%%%%%%%%%%%%%%%%%%%%%%%%%%%%%%%%%%%%%%%%%%%%%%%%%%%%%%%%%%%%%%%
%%%%%%%%%%%%%%%%%%%%%%%%%%%%%%%%%%%%%%%%%%%%%%%%%%%%%%%%%%%%%%%%
%%%%%%%%%%%%%%%%%%%%%%%%%%%%%%%%%%%%%%%%%%%%%%%%%%%%%%%%%%%%%%%%

% \subsection{Deep Neural Network Scaling}
% Here, we demonstrate the efficiency and scalability of \flox{} w.r.t. the size of the model being trained across the network.

% \subsubsection{CHI@Edge Testbed}
% To benchmark our \flox{} framework on edge computing infrastructure, we take advantage of the academic edge test-bed known as CHI@Edge. Part of the Chameleon academic cloud, CHI@Edge is equipped with commonly-used edge computing devices that are commonly used for real-world applications (e.g., Raspberry Pi, NVIDIA Jetson Nano).

% 10 Pis, 4 workstations

% \section{Use Cases}
% \label{sec:usecases}
% \input{sections/6_usecases}

\section{Conclusion \& Future Directions}
\label{sec:conc}
We have presented \flox, the first FL framework to enable on-device and simulated FL in complex hierarchies. 
\flox{} is designed for distributed deployment and adopts novel approaches for decoupling FL control from data movement as well as use of asynchronous communication. 
Support for various launchers allows for simple deployment in different scenarios and scaling from local processes, to parallel execution on a cluster, to wide-area deployment.
\flox{}'s extensible Python implementation and integration with ML frameworks allows it to be easily adopted and customized for various scenarios. 
%meaning it is modular and flexible for a wide range of applications. 
% The inclusion of complex hierarchies means \flox can also be used in real world environments and on devices that are not ideally networked. 
Our experiments demonstrate that \flox{} can scale in multi-node simulations beyond state-of-the-art frameworks.
We also show the ability to reduce data communication requirements by more than 60\% percent through the use of HFL. 
% Additionally, our experiments show only modest
%\matt{@nathaniel, can we put a percent on this? would be nice to say "20\% less time with only 2\% less accuracy or something}\nathaniel{I'm not sure if we want to focus \textit{too} much on model accuracy. This is a result that is so dependent on data distributions, data, and the model which isn't the focus of the work. We should instead probably try to direct the readers to think of the potential for \flox{} to be used to study these problems? That's my initial thought though.} 
% degradation in final model accuracy with increasing hierarchies
Our simulation of AFL demonstrates a nearly 20\% reduction in overall makespan as well as substantially decreased resource idle time. Finally, deployment of HFL in a distributed environment of 11 nodes shows that \flox{} enables effective learning across \workers{} and \aggrs{}.

In future work we will investigate the generalizability of FL algorithms to large scale experiments, the efficient exchange of model weights across different systems, and automated topology configuration to best suit a hierarchical FL workflow.

\section*{Acknowledgements}
This work was supported in part by NSF 2004894 and Laboratory Directed Research
and Development funding from Argonne National Laboratory under U.S. Department of Energy under Contract DE-AC02-06CH11357
and used resources of the Argonne Leadership Computing Facility and SDSC's Expanse Supercomputer.

% \nathaniel{We (may) need to fill this out.}

% \begin{comment}
%% The Appendices part is started with the command \appendix;
%% appendix sections are then done as normal sections
\appendix

\section{Method for Calculating Data Transfer Costs in Hierarchical Topologies}
\label{app:tree_config}

Following the description of the the balanced tree topologies in \autoref{sec:comm_cost_test}.
The data volume can be computed by $2 \cdot E \cdot M$ where 
    $E$ is the number of edges/links/hops in the topology
    and
    $M$ is the size of the model.
By comparison, the total transfer volume in the two-tier case can be estimated by $(E\cdot M) +  (l \cdot h \cdot M)$ where 
    $l$ is the number of leaves (i.e., \workers{}) 
    and 
    $h$ is the height of the topology.
The first term is the initial broadcast and the second term is the response.
From this formulation, it is clear why the total data transfer in two-tier aggregation is larger than hierarchical aggregation.
Further, even without considering transfers across multiple heights, the hierarchical model reduces the data volume handled by the \coordinator{}.

% \section{Appendix title 2}
%% \label{}
% \end{comment}

%% If you have bibdatabase file and want bibtex to generate the
%% bibitems, please use
%%
\balance

\end{document}